\documentclass[sigconf]{acmart}

\AtBeginDocument{%
  }

\usepackage{xspace}
\usepackage{wrapfig}
\usepackage[utf8]{inputenc} %
\usepackage{subfig}
\usepackage[T1]{fontenc}    %
\usepackage{hyperref}       %
\usepackage{url}            %
\usepackage{booktabs}       %
\usepackage{amsfonts}       %
\usepackage{nicefrac}       %
\usepackage{microtype}      %
\usepackage{xcolor}         %
\usepackage{graphicx}
\usepackage{bbm}
\usepackage{paralist}
\usepackage{xspace}
\usepackage{bm}
\usepackage{multirow}
\usepackage{amsmath}
\usepackage{algorithm}
\usepackage{algorithmic} 
\usepackage{enumitem}
\usepackage{pifont}%
\usepackage{mathtools}
\usepackage{makecell}
\usepackage{hyperref}

\usepackage{bbding}
\usepackage{tikz}

\copyrightyear{2024}
\acmYear{2024}
\setcopyright{rightsretained}
\acmConference[KDD '24]{Proceedings of the 30th ACM SIGKDD Conference on Knowledge Discovery and Data Mining}{August 25--29, 2024}{Barcelona, Spain}
\acmBooktitle{Proceedings of the 30th ACM SIGKDD Conference on Knowledge Discovery and Data Mining (KDD '24), August 25--29, 2024, Barcelona, Spain}\acmDOI{10.1145/3637528.3672003}
\acmISBN{979-8-4007-0490-1/24/08}

\makeatletter
\gdef\@copyrightpermission{
 \begin{minipage}{0.3\columnwidth}
  \href{https://creativecommons.org/licenses/by/4.0/}{\includegraphics[width=0.90\textwidth]{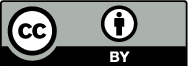}}
 \end{minipage}\hfill
 \begin{minipage}{0.7\columnwidth}
  \href{https://creativecommons.org/licenses/by/4.0/}{This work is licensed under a Creative Commons Attribution International 4.0 License.}
 \end{minipage}
 \vspace{5pt}
}
\makeatother

\newtheorem{problem}{Problem}

\newcommand{\cmark}{\ding{51}}%

\makeatletter
\newcommand\footnoteref[1]{\protected@xdef\@thefnmark{\ref{#1}}\@footnotemark}
\makeatother

\newcommand{\algrule}[1][.5pt]{\par\vskip.5\baselineskip\hrule height #1\par\vskip.5\baselineskip}

\newcommand{\cbit}{\begin{compactitem}}
\newcommand{\ceit}{\end{compactitem}}
\newcommand{\cben}{\begin{compactenum}}
\newcommand{\ceen}{\end{compactenum}}

\newcommand{\beq}{\begin{equation}}
	\newcommand{\eeq}{\end{equation}}

\newcommand{\bit}{\begin{itemize}}
	\newcommand{\eit}{\end{itemize}}
\newcommand{\ben}{\begin{enumerate}}
	\newcommand{\een}{\end{enumerate}}

\newcounter{x}

\newcommand{\lval}{\mathcal{L}_{\text{val}}}
\newcommand{\ltrn}{\mathcal{L}_{\text{trn}}}

\newcommand{\mS}{\mathcal{S}}

\newcommand{\hn}{\widehat{\mathbf{W}}_{\bm{\phi}}}

\newcommand{\bmphi}{\bm{\phi}}

\newcommand{\hnl}{\widehat{\mathbf{W}}_{\bm{\phi}}({\bm{\lambda}})}

\newcommand{\bW}{\mathbf{W}}

\newcommand{\by}{\mathbf{y}}

\newcommand{\bw}{\mathbf{W}}

\newcommand{\R}{\mathbb{R}}

\newcommand{\f}{$f_{\text{val}}$\xspace}

\newcommand{\fval}{f_{\text{val}}}

\newcommand{\bP}{\mathbf{P}}

\newcommand{\bXs}{\mathbf{X}_{\text{test}}}
\newcommand{\bX}{\mathbf{X}}

\newcommand{\vlambda}{{\bm{\lambda}}}
\newcommand{\vLambda}{{\bm{\Lambda}}}
\newcommand{\vepsilon}{\bm{\epsilon}}
\newcommand{\vsigma}{\bm{\sigma}}

\newcommand{\mDt}{\bm{\mathcal{D}}_{\text{train}}}
\newcommand{\mDs}{\bm{\mathcal{D}}_{\text{test}}}
\newcommand{\mD}{\mathcal{D}}

\newcommand{\mO}{\mathcal{O}}
\newcommand{\argmax}{\operatornamewithlimits{argmax}}

\newcommand{\xmark}{\ding{55}}%

\newcommand\crule[3][black]{\textcolor{#1}{\rule{#2}{#3}}}

\definecolor{aliceblue}{rgb}{0.867, 0.917, 0.964}
\definecolor{aliceyellow}{rgb}{0.999, 0.945, 0.796}
\definecolor{alicegray}{rgb}{0.844, 0.867, 0.898}

\newcommand{\lambdaarch}{\bm{\lambda}_{arch}}
\newcommand{\lambdareg}
{\bm{\lambda}_{reg}}
\newcommand{\amask}{\mathbf{A}}

\newcommand{\method}{{\sf H}{\sc y}{\sf P}{\sc er}\xspace}

\begin{document}

\title{Fast Unsupervised Deep Outlier Model Selection with Hypernetworks}

\author{Xueying Ding}
\email{xding2@cs.cmu.edu}
\affiliation{%
  \institution{Carnegie Mellon University}
  \city{Pittsburgh}
  \state{PA}
  \country{USA}
}

\author{Yue Zhao}
\email{yzhao010@usc.edu}
\affiliation{%
  \institution{University of Southern California}
  \city{Los Angeles}
  \state{CA}
  \country{USA}}

\author{Leman Akoglu}
\email{lakoglu@cs.cmu.edu}
\affiliation{%
  \institution{Carnegie Mellon University}
  \city{Pittsburgh}
  \state{PA}
  \country{USA}}

\begin{abstract}
  Deep neural network based Outlier Detection (DOD) has seen a recent surge of attention thanks to the many advances in deep learning. 
In this paper, we consider a
critical-yet-understudied challenge with unsupervised DOD, that is, effective hyperparameter (HP) tuning/model selection. While several prior work report the sensitivity of OD models to HP settings, the issue is ever so critical for the modern DOD models that exhibit a long list of HPs.
We introduce \method for tuning DOD models, tackling two fundamental challenges: (1) validation without supervision (due to lack of labeled outliers), and (2) efficient search of the HP/model space (due to exponential growth in the number of HPs). 
A key idea is to design and train a novel hypernetwork (HN) that maps HPs onto optimal weights of the main DOD model. In turn, \method capitalizes on a \textit{single} HN that can dynamically generate weights for \textit{many} DOD models (corresponding to varying HPs), which offers significant speed-up. In addition, it employs meta-learning on historical OD tasks with labels to train a proxy validation function, %
likewise trained with our proposed HN efficiently. 
Extensive experiments on different OD tasks show that \method achieves competitive performance against 8 baselines with significant efficiency gains.
\end{abstract}




\maketitle

\section{Introduction}
\label{sec:intro}

\textbf{Motivation.} With advances in deep learning, deep neural network (NN) based  outlier detection (DOD) has seen a surge of attention in recent years \cite{pang2021deep,ruff2021unifying}. 
These models, however, inherit many 
hyperparameters (HPs) that can be organized three ways; architectural (e.g. depth, width), regularization (e.g. dropout rate, weight decay), and  optimization HPs (e.g. learning rate). As expected, their performance is highly sensitive to the HP settings \cite{ding2022hyperparameter}. This makes effective HP or model selection critical, yet computationally costly as the model space grows exponentially large in the number of HPs.

Hyperparameter optimization (HPO) can be written as a bilevel problem, where
the optimal parameters $\bw^*$ (i.e. NN weights) on the training set depend on the hyperparameters $\vlambda$.
\beq 
\label{eq:bilevel}
\scalebox{0.90}{
$
\vlambda^* = \arg\min_{\vlambda} \; \lval(\vlambda; \bw^*)
\;\;\;\; s.t. \;\;\;\; 
    \bw^* = \arg\min_{\bw} \; \ltrn(\bw; \vlambda)
$
}
\eeq

\noindent
where $\lval$ and $\ltrn$ denote the validation and training losses, respectively.
There is a body of literature on HPO for supervised settings \cite{bergstra2012random,li2017hyperband,shahriari2015taking}, as well as for OD that use labeled outliers for validation \cite{li2020autood,li2020pyodds,lai2020tods}.
While supervised model selection leverages $\lval$, unsupervised OD posits a unique challenge: it  does not exhibit %
labeled hold-out data to evaluate $\lval$.
It is unreliable to employ the same loss $\ltrn$ as $\lval$ %
as models with minimum training loss do not necessarily associate with accurate detection \cite{ding2022hyperparameter}. 

\begin{figure}[!t]
\hspace{-0.05in}\includegraphics[width=0.485 \textwidth]{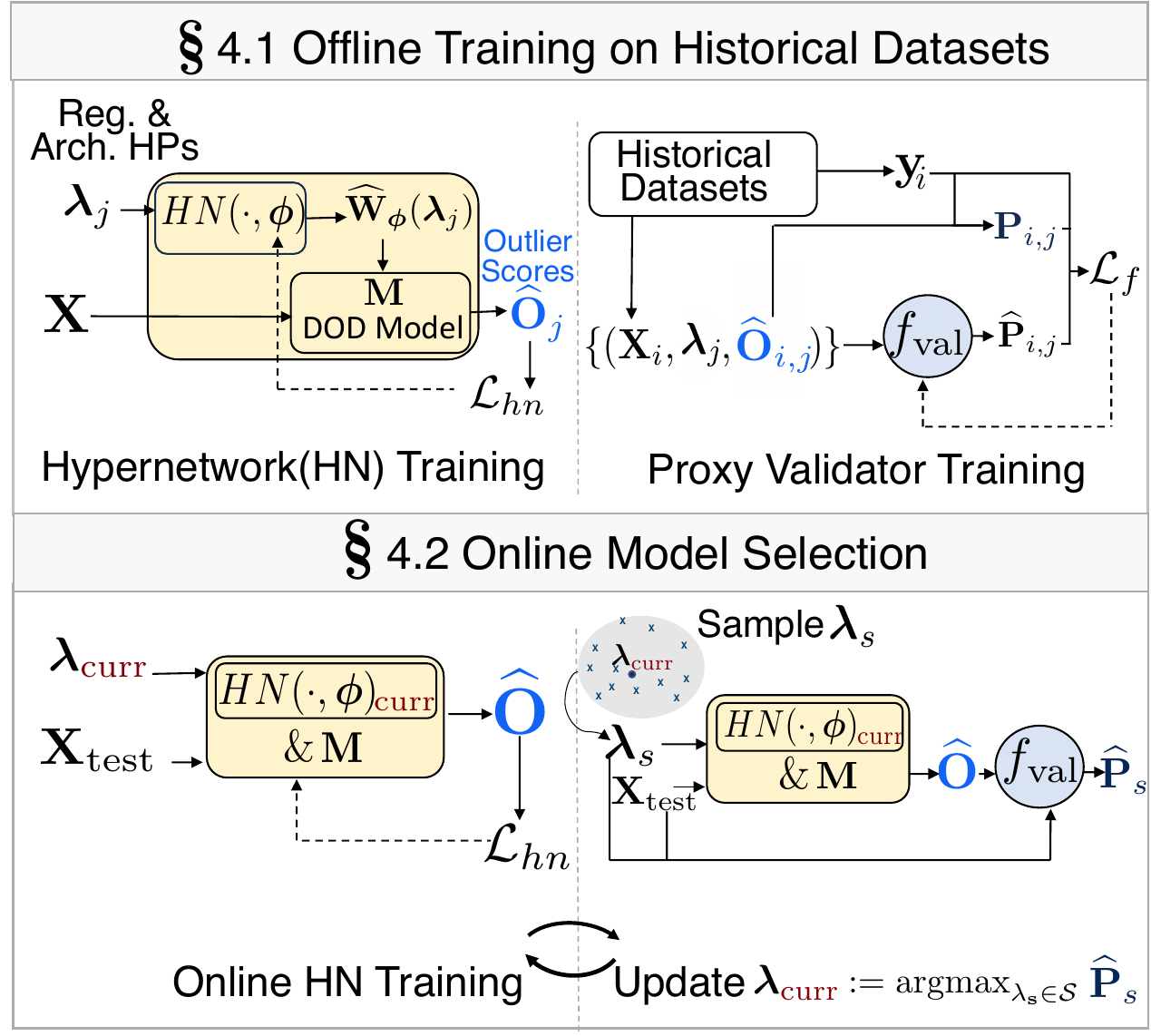}
         \vspace{-0.2in}
        \caption{\method framework illustrated. (top) Offline meta-training of \f (depicted in \crule[aliceblue]{0.2cm}{0.2cm}) on historical datasets for proxy validation 
        (\S \ref{subsubsec:offline}); (bottom) Online model selection on a new dataset (\S \ref{subsubsec:online}). We accelerate both meta-training and model selection using hypernetworks (HN) (depicted in \crule[aliceyellow]{0.2cm}{0.2cm}; \S \ref{ssec:hn}).
        }\label{fig:flow}
        \vspace{-0.2in}
\end{figure}

\textbf{Prior work.~}
Earlier work have proposed {intrinsic} measures for unsupervised model evaluation, based on input data and output (outlier scores) characteristics \cite{journals/corr/Goix16,MarquesCZS15}, using internal consensus among various models \cite{DuanMSWBLH20,lin2020infogan}, as well as properties of the learned weights \cite{martin2021predicting}.  As recent meta-analyses have shown, such intrinsic measures are quite noisy; only slightly and often no better than random \cite{ma2021large}. Moreover, they suffer from the exponential compute cost in large HP spaces %
as they require training numerous candidate models for evaluation. %
{More recent solutions leverage meta-learning by selecting a model for a new dataset based on similar historical datasets \cite{zhao2021automatic,zhao2022toward} . %
They are likewise challenged computationally for large HP spaces and cannot handle any continuous HPs. Our proposed \method also leverages meta-learning, while it is more efficient with the help of hypernetworks as well as more effective by handling continuous HPs with a better-designed proxy validation function. 

\textbf{Present Work.} We introduce \method for unsupervised and efficient hyperparameter tuning for deep-NN based OD. \method tackles both of two key challenges with unsupervised DOD model selection: \textbf{(Ch1)} lack of supervision, and \textbf{(Ch2)} scalability as tempered by the cost of training numerous candidate models.
For \textbf{Ch1}, we employ a meta-learning approach, where we train a proxy validation function, $\fval$, that maps the HPs $\vlambda$, input data, and output outlier scores of DOD models onto corresponding detection performance on historical tasks, as illustrated 
in Fig. \ref{fig:flow} (top, right). 
Since meta-learning builds on past experience (historical datasets) {\em with} labels, the performance of various models can be evaluated.

{Having substituted $\lval$ with meta-trained \f, one can adopt existing %
supervised HPO solutions \cite{feurer2019hyperparameter} toward model selection for a given/new dataset {\em without} labels. However, most of those are susceptible to the scalability challenge (\textbf{Ch2}), as they train each candidate model (with varying $\vlambda$) independently from scratch. To address scalability, and bypass the expensive process of fully training each candidate separately, we leverage \textit{hypernetworks} (HN). This idea is inspired by the self-tuning networks (STN) \cite{MacKayVLDG19}, which estimate the ``{best-response}'' function that maps HPs onto optimal weights through a parameterized hypernetwork (HN), i.e. $\hn(\vlambda) \approx \bW^*$. A single auxiliary HN model 
 can 
 generate the weights of the main DOD model with varying HPs. In essence, it learns how the model weights should change or \textit{respond to} the changes in the HPs (hence the name, best-response), illustrated in Fig. \ref{fig:flow} (top, left).  As a key contribution, we go beyond just the regularization HPs (e.g., dropout rate) that STN \cite{MacKayVLDG19} considered, but propose a novel HN model that can also respond to the \textit{architectural} HPs; including depth and width for DOD models with fully-connected layers.

When a new test dataset (\textit{without} labels) arrives, \method jointly optimizes the HPs $\vlambda$ and the HN parameters ${\bm{\phi}}$ in an alternating fashion, as shown in Fig. \ref{fig:flow} (bottom). Over iterations, it alternates between (1) \textbf{HN training} that updates ${\bm{\phi}}$
to approximate the best-response in a local neighborhood  around the current hyperparameters via $\ltrn$, and then (2) \textbf{HP optimization} that
 updates $\vlambda$ in a gradient-free fashion 
by estimating detection performance through $\fval$ 
of a large set %
of candidate $\vlambda$'s sampled from the same neighborhood,   
using the corresponding HN-generated weights efficiently.  

{\bf Contributions.~}
 \method 
addresses the model selection problem for \textit{unsupervised} deep-NN based outlier detection (DOD),  
applicable to any DOD model, and is \textit{efficient} in the face of the large continuous HP space, tackling both \textbf{Ch1} and \textbf{Ch2}. \method's notable efficiency is  
thanks to our proposed hypernetwork (HN) model that generates DOD model parameters (i.e. NN weights) in response to changes in the HPs associated with regularization as well as NN architecture---in effect, employing a single HN that can act like many DOD models. Further, it offers unsupervised tuning thanks to a proxy validation function trained via meta-learning on historical tasks, which also benefits from the efficiency of our HN.

We compare \method against 8 baselines ranging from simple to state-of-the-art (SOTA) through extensive experiments on 35 benchmark datasets using autoencoder based DOD.  
\method offers the best performance-runtime trade-off, leading to statistically better detection than most baselines; e.g., 2$\times$ $\uparrow$ over default HP in PyOD \cite{zhao2019pyod}), and 4$\times$ speed-up against the SOTA  approach ELECT \cite{zhao2022toward}.

\textbf{Accessibility and Reproducibility}.
We open-source all code and datasets at  \url{https://github.com/xyvivian/HYPER}.

\section{Problem and Preliminaries}
\label{sec:prelim}

The 
sensitivity of outlier detectors to the choice of their hyperparameters (HPs) is well documented \cite{journals/sigkdd/AggarwalS15,CamposZSCMSAH16,ma2021large}.
Deep-NN based OD models are no exception, 
if not even more vulnerable to HP configuration  \cite{ding2022hyperparameter}, as they exhibit a long list of HPs; architectural, regularization and optimization HPs. %
In fact, it would not be an overstatement to point to unsupervised outlier model selection as the primary obstacle to unlocking the ground-breaking potential of deep-NNs for OD.
This is exactly the problem we consider.

\begin{problem}
[Unsupervised Deep Outlier Model Selection (UDOMS)]
\textit{\em Given} a new input dataset (i.e., detection task) $\mDs = (\bXs, \emptyset)$  without any labels, and a deep-NN based OD model $M$;
\textit{\em Output} model parameters corresponding to a selected hyperparameter/model configuration $\vlambda \in \vLambda$ (where $\vLambda$ is the model space)
to employ on $\bXs$ to maximize $M$'s detection performance.
\end{problem}

\noindent
\textbf{Key Challenges:~} Our work addresses two key challenges that arise when  tuning deep neural network models for outlier detection: 
(\textbf{Ch1}) \textit{Validation without supervision} \& 
(\textbf{Ch2}) \textit{Large HP/model space}.

First, unsupervised OD does not exhibit any labels and therefore model selection via validating detection performance on labeled hold-out data is not possible. While model parameters can be estimated end-to-end through unsupervised training losses, such as reconstruction error or one-class losses, one cannot reliably use the same loss as the validation loss; in fact, low error could easily associate with poor detection since most DOD models use point-wise errors as their outlier scores.   

Second, model tuning for the modern OD techniques based on deep-NNs with many HPs is a much larger scale ball-game than that for their shallow counterparts with only 1-2 HPs. This is both due to their ($i$) large number of HPs and also ($ii$) longer training time they typically demand. In other words, the model space that is exponential in the number of HPs and the costly training of individual models necessitate efficient strategies for search.

\section{Fast and Unsupervised: Key Building Blocks of \method}
\label{sec:blocks}
To address the above challenges in UDOMS, 
we propose two primary building blocks for \method: (1) hypernetworks for fast model weight prediction (\S \ref{ssec:hn}) and (2) proxy validator \f that transfers supervision to evaluate model performance on a new dataset without labels (\S \ref{subsec:f_overview}). In \S \ref{sec:full}, we describe how to put together these building blocks toward fast and unsupervised deep OD model selection.

\subsection{Hypernetwork: Train One, Get Many \label{ssec:hn}}

To tackle the challenge of model-building efficiency, we propose a version of hypernetworks (HN) that can efficiently train DOD models with different hyperparameter configurations.
 A hypernetwork (HN) is a network generating weights (i.e. parameters) for another 
 network (in our case, the DOD model) \cite{HaDL17}. Our input to HN is $\vlambda \in \vLambda$, which is one HP configuration and breaks down into two components as $\vlambda = [\lambdareg, \lambdaarch]$, corresponding to regularization HPs (e.g. dropout, weight decay) and architectural HPs (number of layers and width of each layer). Parameterized by $\phi$,
the HN maps a specific hyperparameter configuration 
$\vlambda_j$ to DOD model weights $\hn(\vlambda_j) := HN( \vlambda_j; \phi)$, which parameterize the DOD model for hyperparameter configuration $\vlambda_j$.

We propose 
changes
to the HN \cite{HaDL17}, such that (1) the output $\hn$ can adjust to different architectural shapes, (2) HN can output sufficiently diverse weights in response to varying  $\vlambda$ inputs, and (3) HN training is more efficient than training individual DOD models.

\textbf{Architecture Masking.} To allow the HN output to adapt to various architectures,
we let the output $\hn$'s size be equal to size of the largest architecture in model space $\vLambda$. Then for each  $\lambdaarch$, we build a corresponding architecture masking and feed the masked-version of $\hn$ to the DOD model. In other words, the output $\hn$ handles all smaller architectures by properly padding zeros.

\begin{figure}[h!]
  \centering
  \begin{minipage}[b]{0.42\textwidth}  
    \centering
    \includegraphics[width=0.9\textwidth]{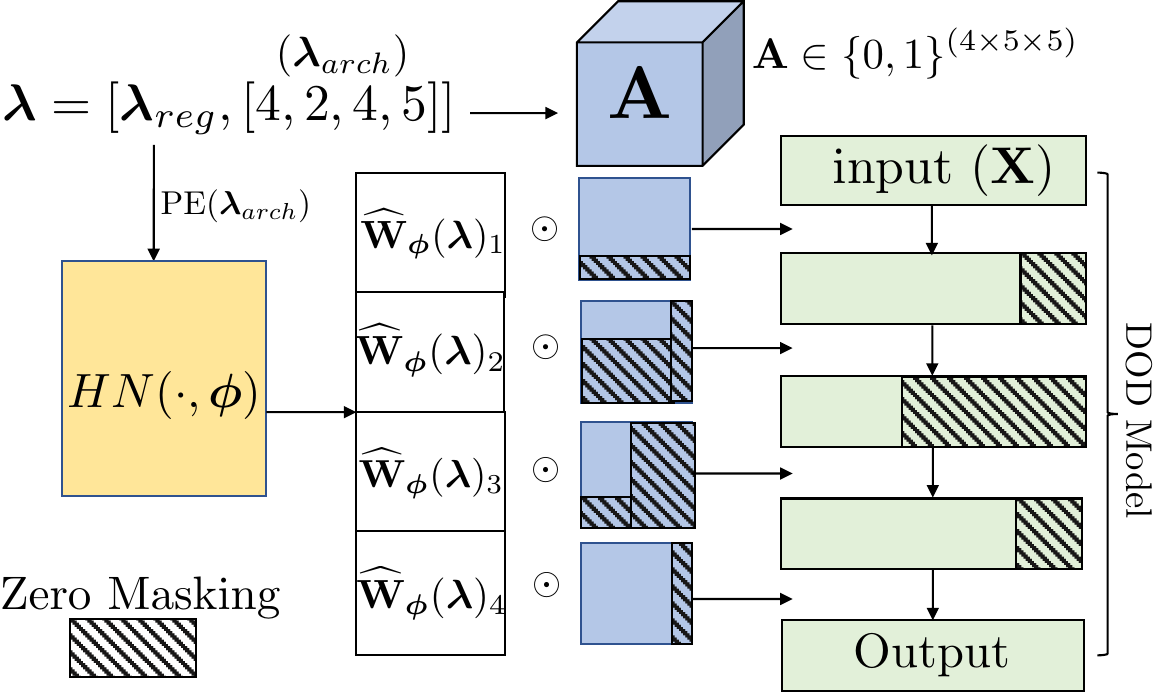}
    \vspace{0.15in}
  \end{minipage}
  \begin{minipage}[b]{0.42\textwidth}
    \centering
    \includegraphics[width=0.9\textwidth]{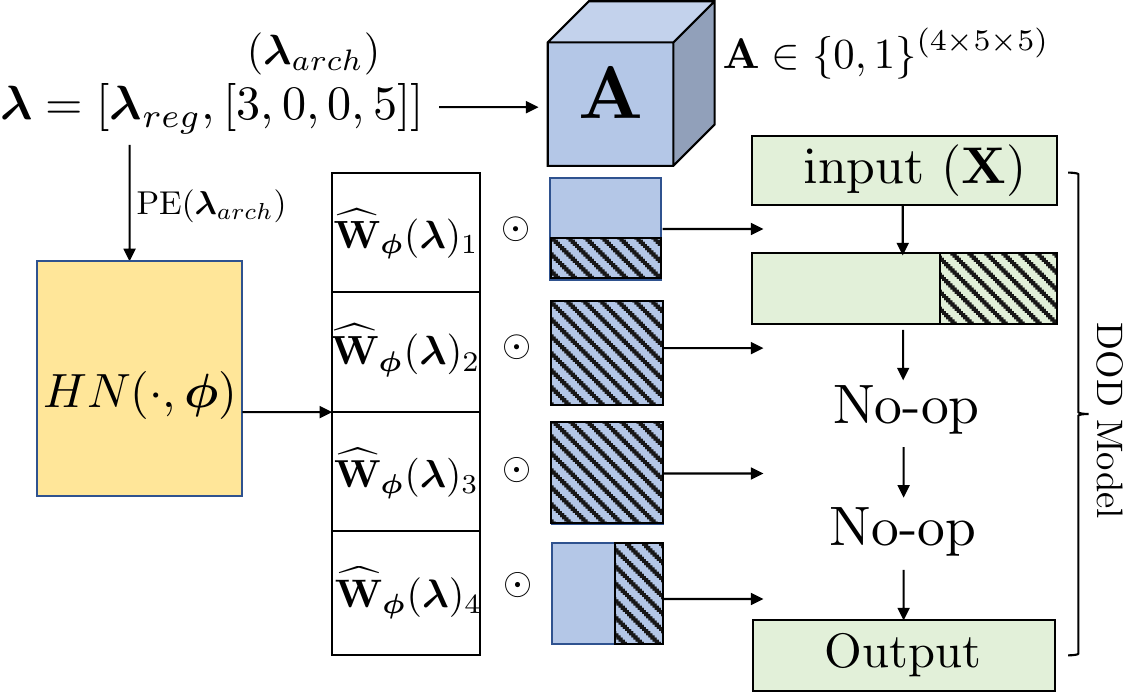}
  \end{minipage}
  \vspace{-0.05in}
  \caption{Illustration of the proposed HN. (Top) HN generates weights for a 4-layer AE, with layer widths equal to $[4,2,4,5]$. Weights $\hn$ is fed into the DOD model, while hidden layers' dimensions are shrunk by the masking $\mathbf{A}$. (Bottom) HN generates weights for a 2-layer AE, with layer widths equal to $[3,5]$. $\vlambda_{arch}$ is padded as $[3,0,0,5]$, and the architecture masking at the second and third layer are set to all zeros. When $\hn$ is fed into the DOD model, zero masking enables the "No Operation" (No-op), in effect shrinking the DOD model from $4$ layers to $2$ layers.}
  \label{fig:HN_example}
  \vspace{-0.27in}
\end{figure}

Taking DOD models built upon MLPs as an example (see Fig. \ref{fig:HN_example}), we make HN output $\hn \in \mathbbm{R}^ {(D \times W \times W)}$, where $D$ and $W$ denote the maximum depth and maximum layer width from $\vLambda$. Assume $\lambdaarch$ contains the abstraction of a smaller architecture; e.g., $L$ layers with corresponding width values $\{W_1, W_2 \ldots, W_L \}$ all less than or equal to $W$. The architectural HP $\vlambda_{arch} \in \mathbbm{N}^{D}$ is defined as:
\begin{align}
\scalebox{0.97}{
$\lambda_{\text{arch}} = [W_1, W_2, \ldots, W_{\lfloor L/2 \rfloor}, \underbracket{{0,\ldots,0}}_{\text{\textcolor{black}{\scriptsize{$(D-L)$ zeros}}}}, W_{\lfloor L/2 \rfloor +1}, \ldots, W_{(L-1)}, W_L] \;.$
}
\nonumber
\end{align}
Then, we construct the architecture masking $\amask \in \{0,1\} ^{(D \times W \times W)}$:

\begin{equation}
 \begin{cases}
    \amask_{[l, 0: \lambdaarch[0],:]} = 1 &\text{, if } l = 0 \\
    \amask_{[l, 0: \lambdaarch[l],0: \lambdaarch[l-z]]} = 1 & \text{, otherwise }  
\end{cases}
\end{equation}
where $\lambdaarch[l-z]$ is the last non-zero entry in $\lambdaarch[0:l]$. (e.g., for $\lambdaarch = [5,3,0,0,3]$ and $l = 4$, the last nonzero entry is $\lambdaarch[1]$ where $z = 3$.)
Then, $l$'th layer weights  %
are multiplied by masking as $\amask_{[l,:,:]} \odot \widehat{\mathbf{W}}_{\bm{\phi},l}$, where non-zero entries are of shrunk dimensions. If $\amask_{[l,:,:]}$ contains only zeros, layer weights become all zeros, representing a "No operation" and is ignored in the DOD model. 

We find that this masking works well with linear autoencoders with a "hourglass" structure, in which case the maximum width $W$ is the input dimension.
For convolutional networks, even though we can tune depths and channels, we can also include kernel sizes and dilation rate by properly padding $\hn$ with zeros \cite{wang2021mergenas}.  We make HN output $\hn \in \mathbbm{R}^{(D \times M_{ch} \times M_{ch}  \times M_{k} \times M_{k})}$, where $D$, $M_{ch}$, $M_{k}$ represent maximum number of layers, channels, and kernel size specified in $\vLambda$, respectively.

Assume $\lambdaarch$ contains the abstraction of a smaller architectures, e.g. $L$ layers with corresponding channel values $\{M_{c1}, M_{c2}, ... , M_{cL} \}$ all less than or equal to $M_{ch}$, and $\{K_1, K_2, ..., K_L\}$ are less than or equal to $M_{k}$. Then, the $\lambdaarch \in \mathbbm{N}^{2D}$ is given as:
\begin{align}
\nonumber
\vlambda_{arch} = [ M_{c1}, K_1, M_{c2}, K_2, \ldots, M_{c\lfloor L/2 \rfloor}, K_{\lfloor L/2 \rfloor} ,\underbracket{{0,\ldots,0}}_{\text{\textcolor{black}{$2(D$$-$$L)$} zeros }},\\ 
M_{c\lfloor L/2 \rfloor +1}, K_{\lfloor L/2 \rfloor +1}, \ldots, M_{cL}, K_L]
\end{align}
The architecture masking $\amask \in \{0,1\}^{(D \times M_{ch} \times M_{ch}  \times M_{k} \times M_{k})}$ is constructed as the following: 

\begin{equation}
\scalebox{0.9}{$
 \begin{cases}
    \amask_{[l, 0: \lambdaarch[2\times l],:,\lfloor \frac{M_{ch}}{2}\rfloor - \lfloor \frac{\lambdaarch[2\times l + 1]}{2}\rfloor: \lfloor \frac{M_{ch}}{2}\rfloor + \lfloor \frac{\lambdaarch[2\times l + 1]}{2}\rfloor]} = 1  \text{, if } l = 0 \\
    \amask_{[l, 0: \lambdaarch[2\times l],0: \lambdaarch[2\times(l-z)],\lfloor \frac{M_{ch}}{2}\rfloor - \lfloor \frac{\lambdaarch[2\times l + 1]}{2}\rfloor: \lfloor \frac{M_{ch}}{2}\rfloor + \lfloor \frac{\lambdaarch[2\times l + 1]}{2}\rfloor]} \\ \qquad\qquad\qquad \qquad\qquad\qquad \qquad\qquad\qquad \qquad\qquad= 1 
    \text{, otherwise }  
\end{cases}
$}
\end{equation}

Again, $\lambdaarch[2\times (l - z) ]$ is the last entry corresponding to the non-zero input channel in $\lambdaarch[2\times l]$. 
Similar to the linear operation, at layer $l$, if $\lambdaarch[2\times l]$ is all zero, then the resulting $\amask_{[l,:,:,:,:]}$ would contain only zeros and represent a "No-op" in the DOD model. 
Otherwise, assume we want obtain a smaller kernel size, $K_l \leq M_{k}$ at layer $l$, the corresponding $\amask_{[l,:,:,:,:]}$ pads zeros around the size $M_{ch} \times k \times k$ center. 
The masked  weights $\amask_{[l,:,:,:,:]} \odot \widehat{\mathbf{W}}_{\bm{\phi},l}$ are equivalent to obtaining smaller-size kernel weights. Notice that, when kernel sizes are different, the output of the layer's operation will also differ (smaller kernels would result in larger output size); therefore, we need to guarantee the spatial size by similarity padding zeros around the input of that convolutional layer. The padding is similar to how we construct the architecture masking $\amask$ and similar to the padding approach discussed in \cite{wang2021mergenas} .

\textbf{Diverse Weight Generation.} While HN is a universal function approximator in theory, it may not generalize well to offer good approximations for many unseen architectures, especially given that the number of $\vlambda$'s  during training is limited. When there is only little variation between two inputs, the HN provides more similar weights, since the weights are generated from the same HN where implicit weight sharing occurs. 

\begin{wrapfigure}{r}{4.85cm}
\centering
\vspace{-0.25 in}
\hspace{-0.2in}
{\includegraphics[width=4.85cm]{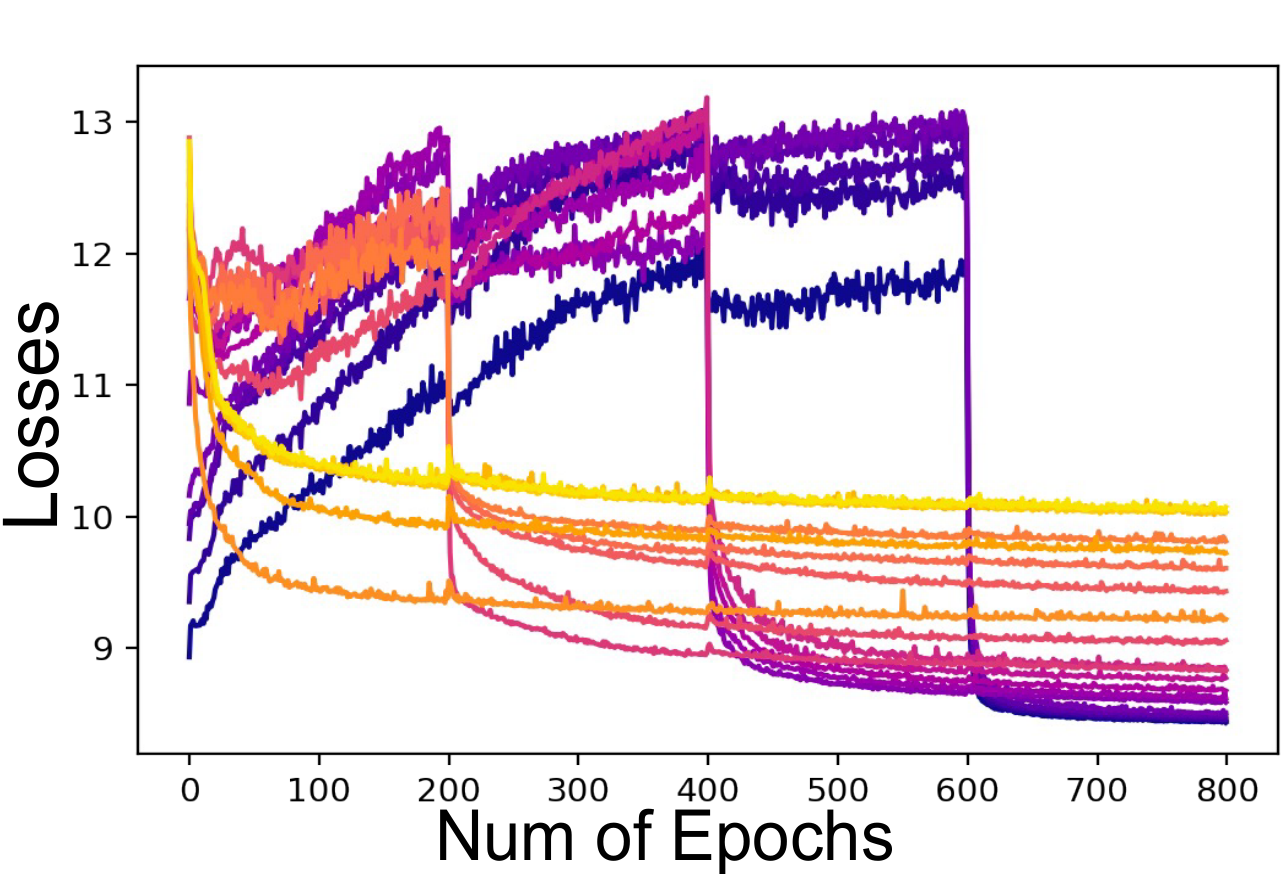} }
\vspace{-0.15 in}
\caption{Loss of individual models during scheduled training. Lighter colors depict loss curves of deeper architectures, which enter training early. Over epochs loss is minimized for all models collectively.} \label{fig:loss}
\vspace{-0.15 in}
\end{wrapfigure}

We employ two ideas toward enabling the HN to generate more expressive weights in response to changes in $\vlambda_{arch}$. First is to inject more variation within its input space where, instead of directly feeding in $\vlambda_{arch}$, we input the positional encoding of each element in $\vlambda_{arch}$.
Positional encoding \cite{positionalencoding} transforms each scalar element into a vector embedding, 
which encodes more granular information, especially when $\vlambda_{arch}$ contains zeros representing a shallower sub-architecture. Second idea is to employ a scheduled training strategy of the HN as it produces weights for both shallow and deep architectures. During HN training, we train with $\vlambda$ associated with deeper architectures first, and later, $\vlambda$ for shallower architectures are trained jointly with deeper architectures. Our scheduled training alleviates the problem of imbalanced weight sharing, where weights associated with shallower layers are updated more frequently as those are used by more number of architectures.  {Fig. \ref{fig:loss}} illustrates 
how the training losses change for individual architectures during the HN's scheduled training.

\textbf{Batchwise Training.}
Like other NNs, HN allows for several inputs $\{ \vlambda_j \}_{j=1} ^m$ synchronously and outputs $\{\hn(\vlambda_j) \}_{j=1}^m $. To speed up training, we batch the input at each forward step with a set  of different architectures and  regularization HP configurations. Assuming $\{\vlambda_j\}_{j=1}^m = \{[\vlambda_{arch,j}, \vlambda_{reg,j}] \}_{j=1}^m \subset \vLambda$ are the set of sampled HP configurations from the model space, given training points $\bX$, the HN loss for one pass is calculated as:
\vspace{-0.1in}
\begin{equation}
    \mathcal{L}_{\text{hn}} = \sum_{\mathbf{x}\in \bX} \sum_{j=1}^m \mathcal{L}_{\text{trn}} \Big ( \hn([\vlambda_{arch,j}, \vlambda_{reg,j}]),\mathbf{x} \Big)
\vspace{-0.07in}
\end{equation}

\noindent
where the training loss $\mathcal{L}_{trn}$ is the same loss as that of the DOD model of interest; e.g. reconstruction loss for autoencoder-based OD models, one-class losses \cite{pmlr-v80-ruff18a}, or regularized reconstruction loss \cite{zhou2017rda},  etc. We feed the HN-generated weights (instead of learning the actual weights) as well as the training data $\bX$ to the DOD model $M$, which then outputs the outlier scores $\widehat{\mO}_{j} := M(\bX; \widehat{\mathbf{W}}_{\bm{\phi}}({\bm{\lambda}_j}))$. The outlier scores are used to compute the training loss $\mathcal{L}_{trn}$, as well as in training our proxy validation function as described next. (See Fig. \ref{fig:flow})
During training, the gradients can further propagate through the generated weights to update the HN parameters $\phi$.

In summary, our HN mimics fast DOD model building across different HP configurations. This offers two advantages: ($i$) training many different HPs jointly in meta-training and ($ii$) fast DOD model parameter generation during online model search. Notably, our HN can tune a wider range of HPs including model architecture, and as shown in \S \ref{sec:ablation}, provides superior results to only tuning $\vlambda_{reg}$.

\subsection{Validation without Supervision via Meta-learning}
\label{subsec:f_overview}

Given the lack of ground truth labels on a new dataset, model selection via supervision is not feasible. Instead,
we consider transferring supervision from historical datasets through meta-learning, enabling model performance evaluation on the new dataset.
Meta-learning uses a collection of historical  tasks $\mDt = \{\mD_1,\ldots,\mD_N\}$ that
contain ground-truth labels, i.e. $\{\mD_i = (\bX_i,\by_i)\}_{i=1}^N$. For many OD settings, such historical datasets are obtainable. For example, for a new disease diagnosis, many previous medical records and medical images are available for normal vs. abnormal health conditions, which can be utilized as historical datasets.  

Given a DOD algorithm $M$ for UDOMS, 
we let $M_j$ denote the DOD model with a specific
HP setting $\vlambda_j$ from the  set 
$\{\vlambda_1, \ldots, \vlambda_m \} \subset \vLambda$. \method uses $\mDt$ to compute two quantities. First, we  obtain \textit{historical outlier scores} of each $M_j$ with HP setting $\vlambda_j$ 
    on each $\mD_i = (\bX_i,\by_i) \in \mDt$. %
    Let $\mO_{i,j} := M_j(\bX_i,\mathbf{W}({\bm{\lambda}_j}))$ denote the output outlier scores of $M_j$ trained with a specific HP configuration $\vlambda_j$ for the data points $\bX_i$ in  dataset $\mD_i$, where $\mathbf{W}({\bm{\lambda}_j})$ are $M_j$'s estimated (i.e. HN-generated) weights for   ${\bm{\lambda}_j}$. Second, we can calculate the \textit{historical performance matrix} $\bP\in \R^{N\times m}$,   
    where $\bP_{i,j}:=\text{perf}(\mO_{i,j}, \by_i)$ denotes $M_j$'s detection performance (e.g. AUROC)
    on dataset $\mD_i$, evaluated based on $\by_i$.

With the historical outlier scores and the performance matrix in hand, we train a proxy validator called \f, which provides us with performance estimation $\widehat{P}$ when we encounter a new dataset and no label is given. As shown in Fig. \ref{fig:flow} (right), the high-level idea of \f is to learn a mapping from 
 data and model characteristics (e.g., distribution of outlier scores) to the corresponding OD performance across $N$ historical datasets and $m$ models (with different HP configurations).
Since it is costly to train all these OD models individually on all datasets from scratch, we utilize our previously proposed HN only \textit{once per dataset} across 
$m$ different HP configurations, which generates the weights and outlier scores for all models. With the hypernetwork and meta-learning from historical dataset introduced, we present the full framework of \method in \S \ref{sec:full} and the training details of \f in \S \ref{subsubsec:offline}.

\section{\method Framework for UDOMS}
\label{sec:full}
\method consists of two phases (see Fig. \ref{fig:flow}): (\S \ref{subsubsec:offline}) offline meta-training 
over the historical datasets, and (\S \ref{subsubsec:online}) online model selection for a given test dataset. 
In the offline phase,
we train the proxy validator $\fval$, %
which allows us to 
predict model performance on the test dataset without relying on labels.
During online model selection, we alternate between training our HN to efficiently generate model weights for varying HPs around a local neighborhood, and refining the best HPs at the current iteration based on $\fval$'s predictions for many locally sampled HPs. We present the details as follows.

\subsection{Offline Training on Historical Datasets}
\label{subsubsec:offline}

In \method, we train a proxy validator \f across historical datasets, so that we can predict the performance of an OD model on the test dataset. 
\f maps HP configuration ($\vlambda$), data embedding, and model embedding onto the corresponding model performance across historical datasets.
The goal is to predict detection performance solely based on the characteristics of the input data and the trained model, along with the HP values.
We create the data embedding and model embedding as described below.

\textbf{Data Embeddings.~} 
Existing work \cite{zhao2021automatic} captures the data characteristics of an OD dataset via extracting meta-features, such as the number of samples and features, to describe a dataset. 
Although simple and intuitive, meta-features primarily focus on general data characteristics with a heuristic extraction process, and are insufficient in model selection \cite{zhao2022toward}. 
In this work, we design a principled approach to capture dataset characteristics. 
First, the datasets may have different feature and sample sizes, which makes it  challenging to learn dataset embeddings. To address this, we employ feature hashing \cite{weinberger2009feature}, $\psi(\cdot)$, to project each dataset to a $k$-dimensional unified feature space. To ensure sufficient
expressiveness, the projection dimension should not be too small ($k$ = 256 in our experiments).
Subsequently, we train a cross-dataset feature extractor $h(\cdot)$, a fully connected neural network, trained with historical datasets' labels, to learn the mapping from hashed samples to the corresponding outlier labels. i.e. $h: \psi(\bX_i) \mapsto \by_i$ for the $i$-th dataset. Training over datasets, the latent representations by $h(\cdot)$ are expected to capture the outlying characteristics of datasets.  Finally, we use max-pooling 
to aggregate sample-wise representations into dataset-wise embeddings, denoted by $\text{pool}\{h(\psi(\textcolor{black}{\bX_i}))\}$.

\textbf{Model Embeddings.~} 
In addition to data embeddings, we need model embeddings that change along with the varying hyperparameter settings to train an effective proxy validator. 
Here we use the historical outlier scores and historical performance matrix, as presented in \S\ref{subsec:f_overview}, to learn  
a neural network $g(\cdot)$ that generates the mapping from the outlier scores onto   detection performance, i.e.
$g: \mO_{i,j} \mapsto \bP_{i,j}$. To handle size variability of outlier scores (due to sample size differences across datasets) as well as to remain agnostic to permutations of outlier scores within a dataset, we employ the DeepSet architecture \cite{zaheer2017deep} for $g(\cdot)$, and use the pooling layer's output as the model embedding, denoted by $\text{pool}\{g(\textcolor{black}{\mO_{i,j}})\}$.

\textbf{Effective and Efficient \f Training.~} By incorporating the aforementioned components, we propose the proxy validator \f, which tries to learn the following mapping:
\begin{align}
    f_{\text{val}} \;:\;
\underbracket{\textcolor{black}{\vlambda_j}}_{\textcolor{black}{\text{HPs}}}, \;
    \underbracket{\text{pool}\{h(\psi(\textcolor{black}{\bX_i}))\}}_{\text{\textcolor{black}{data} embed.}}, \;
    \underbracket{\text{pool}\{g(\textcolor{black}{\mO_{i,j}})\}}_{\text{\textcolor{black}{model} embed.}} \; \mapsto \; \bP_{i,j} \nonumber \\ 
    i\in \{1,\ldots,N\}, \; j\in \{1,\ldots,m\}
    \label{eq:f_train}
\end{align}
\vspace{-0.2in}

We train 
\f with lightGBM \cite{ke2017lightgbm} (one may use any regressor),  
 across $N$ historical datasets and $m$ models with varying HP configurations. The loss function is the squared error between prediction $\mathbf{\widehat{P}}$ and $\bP$: $\mathcal{L}_{f} = \sum_{i=1}^N \sum_{j=1}^m \| \mathbf{\widehat{P}}_{i,j}- \bP_{i,j} \|^2$.

By considering both data and model embeddings in Eq. (\ref{eq:f_train}), \f  predicts performance  more effectively compared to existing works that solely rely on HP values and dataset meta-features \cite{zhao2021automatic}. 

Notice that obtaining the model embeddings in Eq. (\ref{eq:f_train}) that rely on the outlier scores $\mathcal{O}_{i,j}$  
requires training the DOD model for each dataset $\mD_i$ and each HP configuration $\bm{\lambda}_j$, which can be computationally expensive. To speed up this process for meta-training,
we use HN-generated weights rather than training these individual DOD models from scratch,  to obtain  %
$\widehat{\mO}_{i,j} := M_j(\bX_i; \widehat{\mathbf{W}}^{(i)}_{\bm{\phi}}({\bm{\lambda}_j}))$, where $\widehat{\mathbf{W}}^{(i)}_{\bm{\phi}}({\bm{\lambda}_j})$ denotes model $M_j$'s weights for HP configuration $\vlambda_j$, as generated by our HN trained on $\mathcal{D}_i$. 

At test time, the proxy validator \f provides performance evaluation by taking in the new dataset's embedding and a DOD model's embedding with a HP configuration, \textit{without} requiring the ground truth labels. In this way, we are able to leverage the benefits of meta-learning from historical datasets, and  estimate the performance of a specific DOD model on a new dataset, when the new dataset contains no labels.

\subsection{Online Model Selection}
\label{subsubsec:online}

\textbf{Model selection via proxy validator.~} 
With our meta-trained $\fval$ at hand, given a test dataset $\bXs$, a simple model selection strategy would be to train many DOD models with different randomly sampled HPs on  $\bXs$ to obtain outlier scores, and then select the one with the highest predicted performance by $\fval$.

However, training OD models from scratch for each HP can be computationally expensive.
To speed this up, we also train a HN on $\bXs$ and subsequently obtain the outlier scores $\widehat{\mO}_{\text{test}}$ from the DOD model with HN-generated weights for randomly sampled HPs. We select the best HP configuration according to  Eq. (\ref{eq:selected_model}).

\begin{equation}
    \argmax_{\vlambda \in \vLambda} \; f_{\text{val}}(\bXs, \vlambda, \widehat{\mO}_{\text{test},\lambda})
 \label{eq:selected_model}
\end{equation}

Moreover, we propose to iteratively train our HN over locally selected HPs, since training a ``global'' HN to generate weights across the entire $\vLambda$ and over unseen $\vlambda$ is a challenging task especially for large model spaces \cite{MacKayVLDG19}, which can impact the quality of the generated weights and subsequently affect the overall performance of the selected model. Therefore, we propose two stages of alternate updating. One stage trains the HN according to a neighborhood of sampled HP configurations around the current best HP, while the other stage applies the generated weights from HN to obtain outlier scores, and subsequently find the new best HP configuration through the performance proxy validator.

\renewcommand{\algorithmicrequire}{\textbf{Input:}}
\renewcommand{\algorithmicensure}{\textbf{Output:}}
\renewcommand{\algorithmiccomment}[1]{\hfill$\blacktriangleright$ #1}

\begin{algorithm}[!t]
	\caption{\method: Online Model Selection 
 }
	\label{algo:ours}
 {
	\begin{algorithmic}[1]
		\REQUIRE test dataset $\mDs=(\bXs, \emptyset)$, 
  HN parameters $\bm{\phi}$, HN learning rate $\alpha$, HN loss  $\mathcal{L}_{\text{hn}}(\cdot)$, proxy validator \f, HN (re-)training epochs $T$, 
  validation objective $\mathcal{G}(\cdot)$ in Eq. (\ref{eq:true_objective}), 
  patience $p$
		\ENSURE optimized HP configuration $\vlambda^*$ for the test dataset
		\algrule
            \STATE Initialize 
            $\vlambda_{\text{curr}}$
            and $\vsigma_{\text{curr}}$

		\WHILE%
  {patience criterion $p$ is not met}
  \STATE Sample set  $\mS := \emptyset$
                \FOR{$t=1, \ldots, T$}
                \STATE $\vepsilon_t \sim p(\vepsilon|\bm{\sigma}_{\text{curr}})$ \COMMENT{local sampling range  around $\vlambda_{\text{curr}}$}
                \STATE $\bm{\phi} \leftarrow \bm{\phi} -  \alpha \frac{\partial}{\partial \phi} \mathcal{L}_{\text{hn}}(\vlambda_{\text{curr}}+\vepsilon_t, \widehat{\mathbf{W}}_{\bmphi}(\vlambda_{\text{curr}}+\vepsilon_t))$
                \STATE $\mS := \mS \cup (\vlambda_{\text{curr}}+\vepsilon_t)$
                \COMMENT save locally sampled HPs 
                \ENDFOR

                \STATE $\vlambda_{\text{curr}} \leftarrow \text{argmax}_{\vlambda \in \mS} \;\; \mathcal{G(\vlambda, \vsigma_{\text{curr}};  
                {\bm{\phi}})}
                 $ \COMMENT{Eq. (\ref{eq:true_objective}) } 

                \STATE $\vsigma_{\text{curr}} \leftarrow \text{argmax}_{\vsigma
                } \; \mathcal{G(\vlambda_{\text{curr}}, \vsigma;
                {\bm{\phi}})}
                $

            \ENDWHILE
            \RETURN 
            $\vlambda^* \approx \argmax_{\vlambda \in \mS} \; f_{\text{val}}(\bXs, \vlambda, \widehat{\mO}_{\text{test},{\bm{\lambda}}})$
             \COMMENT{Eq. (\ref{eq:final_model})}

	\end{algorithmic}
 }
\end{algorithm}
\setlength{\textfloatsep}{0.05in}

\textbf{Training local HN iteratively and adaptively.~}
We design \method to jointly optimize the HPs $\vlambda$ and the (\textit{local}) HN parameters ${\bm{\phi}}$ in an alternating fashion; as shown in Fig. \ref{fig:flow} (bottom). Algorithm \ref{algo:ours} provides the step-by-step outline of the  process.
Over iterations, it {alternates} between two stages:

\vspace{-0.05in}

\begin{enumerate}[leftmargin=0.9cm]
\setlength\itemsep{0.005in}
    \item[(S1)] \textbf{HN training} that updates HN parameters ${\bm{\phi}}$ to approximate the best-response in a local neighborhood 
    around the current hyperparameters $\vlambda_{\text{curr}}$ via %
    $\mathcal{L}_{\text{hn}}$ (Lines 4--8), and
    \item[(S2)] \textbf{HP optimization}  that 
 updates $\vlambda_{\text{curr}}$ in a gradient-free fashion by estimating detection performance through $\fval$ of a large set %
 of candidate $\vlambda$'s sampled from the same neighborhood,   
using the corresponding approximate best-response, i.e. the HN-generated weights, $\hnl$ (Line 9). 
\end{enumerate}

\vspace{-0.05in}

To dynamically control the sampling range around $\vlambda_{\text{curr}}$, we use a factorized Gaussian with standard deviation $\vsigma$ to generate local HP perturbations $p(\vepsilon|\vsigma)$. We initialize $\vsigma$ to be a scale factor vector, each value is within $\mathbb{R}^+$, and dynamically change the value of $\vsigma$, which becomes $\vsigma_{\text{curr}}$ to control the radius of sampling neighborhood.
$\vsigma_{\text{curr}}$ is used in (S1) for sampling local HPs and is then updated in (S2) at each iteration (Line 10). 

\textbf{Updating $\vlambda_{\text{curr}}$ and  $\vsigma_{\text{curr}}$}. \method iteratively explores  promising HPs and the corresponding sampling range. To update $\vlambda_{\text{curr}}$ and 
 the sampling factor 
$\vsigma_{\text{curr}}$, we 
maximize:
\vspace{-0.05in}
\begin{align}
\label{eq:objective} 
\underbracket{\mathbb{E}_{\vepsilon \sim p(\vepsilon|\bm{\sigma})}[\fval(\bXs, \vlambda+\vepsilon, \widehat{\mathbf{W}}_{\bmphi}(\vlambda+\vepsilon))]}_{\text{
     update } \vlambda_{\text{curr}} \text{ to a better model/HPs w/ high expected performance}} \\ \nonumber + 
     \underbracket{\tau \; \mathbb{H}(p(\vepsilon|\vsigma))}_{\text{
    sampling range around }\vlambda_{\text{curr}}} \;.
\end{align}
The objective consists of two terms. 
The first term emphasizes selecting the next model/HP configuration with high expected performance, aiming to improve the overall model performance. 
The second term measures the uncertainty of the sampling factor, quantified with Shannon's entropy $\mathbb{H}$. A higher entropy value indicates a less localized sampling, allowing for more exploration.
The objective is to find an HP configuration that can achieve high expected performance, within a 
reasonably large 
 local region to contain a good model, that is also local enough for the HN to be able to effectively learn the best-response. If the sampling factor $\vsigma$ is too small, it limits the exploration of the next HP configuration and training of the HN, potentially missing out on better-performing options. Conversely, if $\vsigma$ is too large, it may lead to inaccuracies in the HN's generated weights, compromising the accuracy of the first term. The balance factor $\tau$ controls the trade-off between the two terms. 

We approximate the expectation term in Eq. (\ref{eq:objective}) by the empirical mean of predicted performances  through $V$ number of sampled perturbations around $\vlambda$.
We define our validation 
objective $\mathcal{G}$ as: 

\vspace{-0.2in}
\begin{equation}
\label{eq:true_objective}
\scalebox{0.925}{
$\mathcal{G}(\vlambda, \vsigma; {\bmphi}) = \frac{1}{V} \sum_{i=1}^{V} \fval(\bXs, \vlambda+\vepsilon_i, \widehat{\mathbf{W}}_{\bmphi}(\vlambda+\vepsilon_i)) + \tau \mathbb{H}(p(\vepsilon|\vsigma))$}
\end{equation}

In each iteration of the HP update, we first fix $\vsigma_{\text{curr}}$ and find the configuration in $\mS$ with the highest value of Eq. (\ref{eq:true_objective}), where we sample $V$ local configurations around %
each $\vlambda \in \mS$, 
i.e., $\vlambda + \vepsilon_i| \vsigma_{\text{curr}}$ for $i \in 1, \ldots, V$. After %
$\vlambda_{\text{curr}}$ is updated, we fix it and update the sampling factor $\vsigma_{\text{curr}}$ also by Eq. (\ref{eq:true_objective}), using $V$ samples  based on each $\vsigma$  from a pre-specified range for each HP. To %
ensure encountering a good HP configuration, 
we set $V$ to be a large number, e.g. 500. We provide details and pre-specified range in Appx. \S\ref{appx:setting_baselines}.

\textbf{Selecting the Best Model/HP $\vlambda^*$}. We employ \f to  choose the best HP $\vlambda^*$ among  all the locally sampled HPs $\mS$ during the last iteration of HN training. 
Note that \method directly uses the HN-generated weights %
for fast computation, without the need to train any model from scratch for evaluation by \f. With the generated weights $\widehat{\mathbf{W}}_{\phi}(\vlambda)$ , the DOD model produces the corresponding outlier scores, denoted as %
$\widehat{\mO}_{\text{test},{\bm{\lambda}}} := M_{\bm{\lambda}}(\bXs; \widehat{\mathbf{W}}_{\bm{\phi}}({\bm{\lambda}}))$,
that allows us to select the best HP configuration within the candidate set by 
\begin{equation}
    \vlambda^* \approx \argmax_{\vlambda \in \mS} \; f_{\text{val}}(\bXs, \vlambda, \widehat{\mO}_{\text{test},{\bm{\lambda}}}) \;.
    \label{eq:final_model}
\vspace{-0.05in}
\end{equation}

\textbf{Initialization and Convergence.} 
We initialize $\vlambda_{\text{curr}}$ and $\vsigma_{\text{curr}}$ with the globally best values across historical datasets.
We consider 
\method as converged if the highest predicted performance by \f does not improve in $p$ consecutive iterations. A larger $p$, referred as  ``patience'', 
requires more iterations to converge yet likely yields better results. Note that $p$ can be decided by cross-validation on historical datasets during meta-training. 
We present an empirical analysis of initialization 
and patience in the experiments.

\section{Experiments}
\label{sec:experiments}
\subsection{Experiment Settings}

\textbf{Benchmark Data.} 
We show \method's effectiveness and efficiency with fully connected AutoEncoder (AE) for DOD on tabular data, using a testbed consisting of 34 benchmark datasets from two different public OD repositories; ODDS \cite{Rayana:2016} and DAMI \cite{Campos2016} (\texttt{Pima} dataset is removed). In addition, we run \method with convolutional AE on MNIST and FashinMNIST datasets. We treat one class as the normal class, while downweighting the ratio of another
class at 10\% as the outlier class. We train and validate \method with
inliers and outliers from classes [0,5] (30 tasks/datasets in total) and evaluate 8 tasks on
[6,9], to avoid data leakage in (meta)training/testing data.

\begin{table}[]

\caption{\method and baselines for time (in mins) and performance comparison with categorization by whether it selects models (2nd column), uses meta-learning (3rd column), and requires model building at the test time (4th column). Overall, \method (with patience $p=3$) achieves the best detection performances (also see Fig. \ref{fig:perf_ae_time} and \ref{fig:perf_ae_barplot}). Compared to the SOTA ELECT, \method has markedly shorter offline and online time.
 } %
 \vspace{-0.1in}
\centering
\scalebox{0.67}{
\begin{tabular}{l|ccc|crr|c}
\toprule
\textbf{Method} & \makecell{\textbf{Model}\\ \textbf{Selection}} & \makecell{\textbf{Meta}\\ \textbf{Learning}} & \makecell{\textbf{Zero} \\ \textbf{shot}} & \makecell{\textbf{Offline} \\ \textbf{Time}} & \makecell{\textbf{Avg. On} \\ \textbf{-line Time}} & \makecell{\textbf{Med. On} \\ \textbf{-line Time}} & \makecell{\textbf{Avg. ROC} \\ { \textbf{Rank}($\downarrow$})} 
\\
\hline
Default           &\xmark                       &\xmark &\cmark                     & N/A                   & 0                 & 0    &               0.5954 \\
Random          &\xmark                       &\xmark       &\cmark              & N/A                   & 0                 & 0                 & 0.5603  \\
MC        &\cmark                       &\xmark    &\xmark                 & N/A                & 215               & 277        & 0.5642  \\
GB     &\cmark                       &\cmark        &\cmark             & 7,461                & 0                    & 0       & 0.4668       \\
ISAC            &\cmark                       &\cmark      &\cmark               & 7,466                & 1                    & 1               & 0.4181  \\
AS              &\cmark                       &\cmark     &\cmark                & 7,465                & 1                    & 1               & 0.5222  \\
MetaOD           &\cmark                       &\cmark    &\cmark                 & 7,525                & 1                    & 1              & 0.3918 \\
ELECT           &\cmark                       &\cmark    &\xmark                 & 7,611                & 59               & 71     & 0.3621       \\
\hline
Ours            &\cmark                       &\cmark      &\xmark               & 1,320                 & 14                & 17  & 0.2954\\     \bottomrule        
\end{tabular}
}
	
	\label{table:baseline} %
\end{table}
\setlength\tabcolsep{6 pt}

\textbf{Baselines.} For tabular dataset, we include 8 baselines for comparison ranging from simple to state-of-the-art (SOTA); Table \ref{table:baseline} provides a conceptual comparison of the baselines. They are organized as (\textit{i}) \textbf{\textit{no model selection}}: \textbf{(1) Default} uses the default HPs used in a popular OD library PyOD \cite{zhao2019pyod}, \textbf{(2) Random} picks an HP/model randomly (we report expected performance); (\textit{ii}): \textbf{\textit{model selection without meta-learning}}: \textbf{(3) MC} \cite{ma2021large} leverages consensus;
and (\textit{iii}) \textbf{\textit{model selection by meta-learning}}: \textbf{(4) Global Best (GB)} selects the best performing model on the historical datasets on average, and SOTA baselines include 
\textbf{(5) ISAC \cite{conf/ecai/KadiogluMST10}},
\textbf{(6) ARGOSMART (AS) \cite{nikolic2013simple}}, 
\textbf{(7) MetaOD} \cite{zhao2021automatic}, 
\textbf{(8) ELECT} \cite{zhao2022toward}. 
Baselines (1), (2), and (4)-(7) are zero-shot that do not require model building during model selection. For image dataset, we proivde \method's performance in comparison to Random (avg. performance across HPs) and Default (configuration used by DeepSVDD \cite{deepsvdd}).

\textbf{Evaluation.} For tabular data, we use 5-fold cross-validation to split the train/test datasets; that is, each time we use 28 datasets as the historical datasets to select models on the remaining 7 datasets. For image data, we use 5-fold cross-validation: 24 (out of 30) datasets as the historical datasets to select models on the remaining 6 (meta-train) datasets.
We use AUROC to measure detection performance, while it can be substituted with any other measure. 
We report the raw ROC as well as the normalized ROC \textit{Rank} of an HP/model, ranging from 0 (the best) to 1 (the worst)---i.e., the lower the better.We use the paired Wilcoxon signed rank test \cite{Groggel00Stats} across all datasets in the testbed to compare two methods. 
Full results in Appx. \S \ref{appx:full_results}. 

\vspace{-0.05in}
\subsection{Experiment Results}

\begin{figure}
\vspace{-0.15in}
\includegraphics[width=6.5cm]{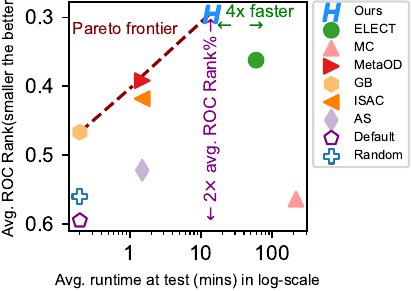}
\vspace{-0.15in}
\caption{Avg. running time (log-scale) vs. avg. model ROC Rank. Meta-learning methods are depicted with solid markers. Pareto frontier (red dashed line) shows the best methods under different time budgets. \method outperforms all with reasonable computational demand.} \label{fig:perf_ae_time}
\vspace{-0.1in}
\end{figure}

\textbf{Tabular.} Fig. \ref{fig:perf_ae_time}
shows that \textbf{\method outperforms all baselines with regard to average ROC Rank 
on the 35 testbed}.In addition, Fig. \ref{fig:perf_ae_barplot} provides the full performance distribution across all datasets and shows that \method is statistically better than all baselines, including
 SOTA meta-learning based ELECT and MetaOD. 
Among the zero-shot baselines, Default and Random  perform significantly poorly while the meta-learning based GB leads to comparably higher performance. Same as previous study \cite{ma2021large}, the internal consensus-based MC can be no better than Random.

\begin{figure}
\centering
\vspace{-0.10in}
{\includegraphics[width=7cm]{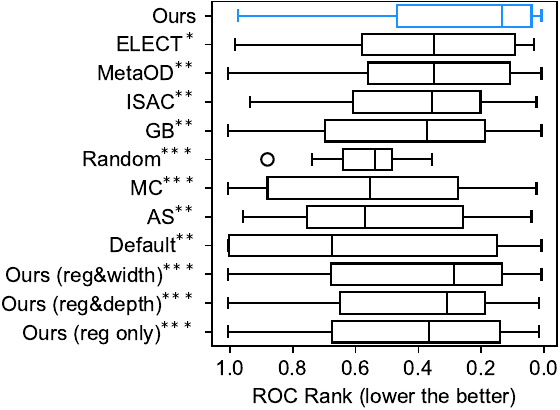} }
\vspace{-0.1in}
\caption{Distribution of ROC Rank across datasets. \method achieves the best performance. 
Bottom three bars depict \method's variants that do not fully tune architecture HPs (for ablation). Paired test results are depicted as significant w/ $^{*}$ at 0.1, $^{**}$ at 0.01, $^{***}$ at 0.001. See $p$-values in Appx. Table \ref{table:ae_pairs}.} 
\label{fig:perf_ae_barplot}
\vspace{0.1in}
\end{figure}

\textbf{HN-powered efficiency enables \method to search more broadly}. Fig. \ref{fig:perf_ae_time} and Appx. Table \ref{table:baseline} show that \method offers significant speed up over the SOTA method ELECT, with an average offline training speed-up of 5.77$\times$ and a model selection speed-up of 4.21$\times$. 
Unlike ELECT, which requires building OD models from scratch during both offline and online phases, \method leverages the HN-generated weights 
to avoid costly model training for each candidate HP.

Meanwhile, \method can also afford a broader range of HP configurations thanks to the lower model building cost by HN. This 
capability contributes to the effectiveness of \method, which brings 7\% avg. ROC Rank $\uparrow$ over ELECT.

\textbf{Meta-learning methods achieve the best performance at different budgets}.
Fig. \ref{fig:perf_ae_time} and Appx. Table \ref{table:baseline} show that the best performers at different time budgets are global best (GB), MetaOD, and \method, which are all on the Pareto frontier. 
In contrast, simple no-model-selection approaches, i.e., Default and Random, are typically the lowest performing methods. 
Specifically, \method achieves a significant 2$\times$ avg. ROC Rank improvement over the default HP in PyOD \cite{zhao2019pyod}), a widely used open-source OD library. 
Although meta-learning entails additional (offline) training time,
 it can be amortized across multiple future tasks in the long run.

 \textbf{Image.} Table \ref{tab:mnist} shows that \method is outperforming both Random selection and Default LeNet AutoEncoder model, as it is able to effectively learn from historical data to tune HPs. In addition, we evaluate \method’s performance across datasets by online training with 5 FashionMNIST tasks, with the same search space of HPs and same offline-training solely on MNIST anomaly detection tasks. Table \ref{tab:fashmnist} shows that \method outperforms Random on average,but not necessarily on all datasets. It may be due to the fact that MNIST (used for meta-learning) and FashionMNIST (test tasks) are from different distributions and share less similarities for effective transfer through meta-learning.

\begin{table}[ht]
\centering
\caption{Test AUROC of \method in comparison to Random and Default HP configurations. Overall, \method shows higher AUROC than Random and Default.}
\vspace{-0.12in}
\label{tab:mnist}
\begin{tabular}{lccc}
\hline
\textbf{Inlier/Outlier Class} & \textbf{Random} & \textbf{Default} & \method \\
\hline
Inlier: 6 Outlier: 9 & 0.8343 & 0.8015 & \textbf{0.9463} \\
Inlier: 9 Outlier: 7 & 0.6358 & 0.6409 & \textbf{0.7761} \\
Inlier: 8 Outlier: 6 & 0.7152 & 0.6358 & \textbf{0.8490} \\
Inlier: 6 Outlier: 8 & 0.7827 & 0.8276 & \textbf{0.8365} \\
Inlier: 6 Outlier: 7 & 0.7787 & 0.7968 & \textbf{0.9179} \\
Inlier: 7 Outlier: 6 & 0.8720 & 0.8330 & \textbf{0.8909} \\
Inlier: 9 Outlier: 6 & 0.7642 & 0.7371 & \textbf{0.9098} \\
Inlier: 7 Outlier: 9 & 0.4073 & \textbf{0.5769} & 0.4613 \\
\hline
\end{tabular}
\end{table}

 \vspace{-0.16in}
\begin{table}[ht]
\centering
\caption{Test AUROC of \method and Random, evaluated on FashionMNIST Datasets.\method, solely trained on MNIST, does not provide competitive performances in all cases.}
\vspace{-0.12in}
\label{tab:fashmnist}
\begin{tabular}{lcc}
\hline
\textbf{Inlier/Outlier Class} & \textbf{Random} & \method \\
\hline
Inlier: T-shirt Outlier: Trouser & 0.6004 & \textbf{0.6132} \\
Inlier: Trouser Outlier: Pullover & 0.8902 & \textbf{0.9878} \\
Inlier: Dress Outlier: Coat  & \textbf{0.8715} & 0.8456 \\
Inlier: Sandal Outlier: Shirt  & \textbf{0.9024} & 0.8752 \\
Inlier: Sneaker Outlier: Bag & \textbf{0.9457} & 0.9033 \\
\hline
\end{tabular}
\end{table}

\subsection{Ablation Studies}
\label{sec:ablation}

\textbf{Benefit of Tuning Architectural HPs via HN}. 
\method tackles the challenging task of accommodating architectural HPs besides regularization HPs. Through ablations, we study the benefit of our novel HN design, as presented in \S \ref{ssec:hn}, which can generate DOD model weights in response to changes in architecture.  
 Bottom three bars of Fig. \ref{fig:perf_ae_barplot} 
show the performances of three \method variants across datasets.  %
The proposed
\method (with median ROC Rank = 0.1349)  outperforms all these variants significantly (with $p$$<$$0.001$), namely, \textit{only tuning regularization and width} (median ROC Rank = 0.2857), \textit{only tuning regularization and depth} (median ROC Rank = 0.3095), and \textit{only tuning regularization} (median ROC Rank = 0.3650). By extending its search for both neural network depth \textit{and} width, \method explores a larger model space that helps find better-performing model configurations.

\begin{figure}
\centering
\vspace{-0.05in}

{\hspace{-0.13in}\subfloat
        {%
			\includegraphics[clip,width=0.25\columnwidth]{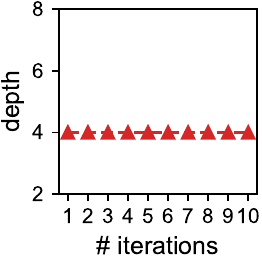}%
	}
	\hspace{-0.02in}
        {%
			\includegraphics[clip,width=0.25\columnwidth]{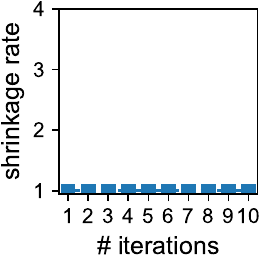}%
	}
	\hspace{0.02in}
        {%
			\includegraphics[clip,width=0.25\columnwidth]{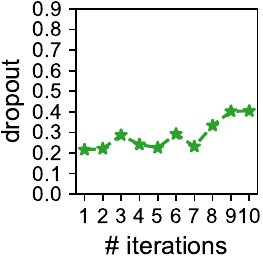}%
	}
        \hspace{-0.02in}
        {%
			\includegraphics[clip,width=0.25\columnwidth]{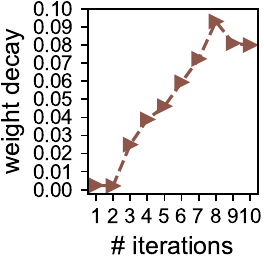}%
	}

        \hspace{-0.15in}
 	\subfloat
        {%
			\includegraphics[clip,width=0.25\columnwidth]{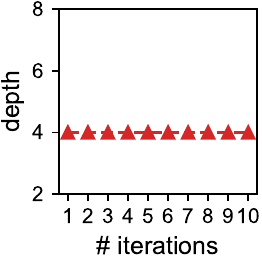}%
	}
	\hspace{-0.02in}
        {%
			\includegraphics[clip,width=0.25\columnwidth]{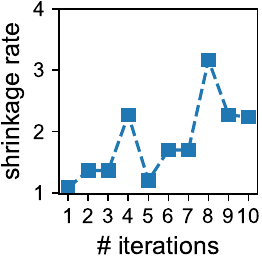}%
	}
	\hspace{0.02in}
        {%
			\includegraphics[clip,width=0.25\columnwidth]{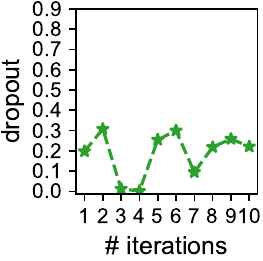}%
	}
        {%
			\includegraphics[clip,width=0.25\columnwidth]{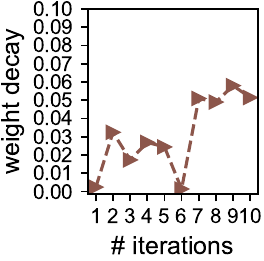}%
	}

}

\vspace{-0.05in}
\caption{Trace of HP changes over iterations on \texttt{spamspace}: (top) tuning regularization HPs only; (bottom) tuning both regularization and architectural HPs (ours). When arch. is fixed, reg. HPs incur more magnitude changes and reach larger values to adjust model complexity. \method tunes complexity more flexibly by also accommodating arch. HPs.} \label{fig:trace}
\vspace{0.1in}
\end{figure}

\textbf{HP Schedules over Iterations}. In Fig. \ref{fig:trace}, we more closely analyze  how HPs change over iterations on \texttt{spamspace}, comparing between (top) only tuning reg. HPs while fixing model depth and width (i.e., shrinkage rate) and (bottom) using \method to tune all HPs including both reg. and architectural HPs. Bottom figures show that %
depth remains fixed at 4, shrinkage rate increases from 1 to 2.25 (i.e., width gets reduced), dropout to 0.2, and weight decay to 0.05---overall model capacity is reduced relative to initialization. 
In contrast, top figures show that, when model depth and width are fixed, regularization HPs compensate more to adjust the model capacity, with a larger dropout rate at 0.4 and larger weight decay at 0.08, achieving ROC rank 0.3227 in contrast to \method's 0.0555. This comparison showcases the merit of \method which adjusts model complexity more flexibly by accommodating a larger model space.

\begin{figure}[!ht]
    \centering
    \subfloat %
    {
    \includegraphics[clip,width=0.45\columnwidth]{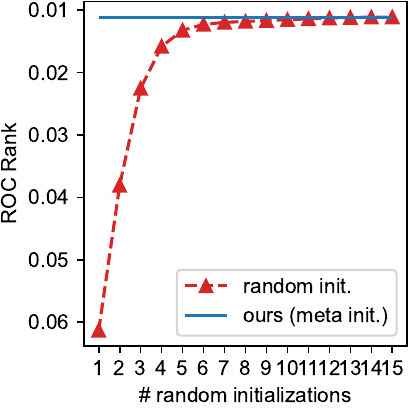}
    }
    \subfloat %
    {
    \includegraphics[clip,width=0.45\columnwidth]{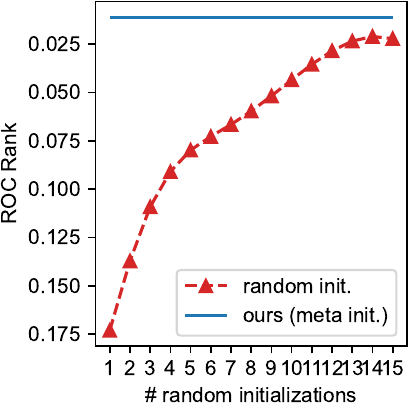}
    }
     \vspace{-0.1in}
     \hspace{0.05in}
    \subfloat %
    {
    \includegraphics[clip,width=0.45\columnwidth]{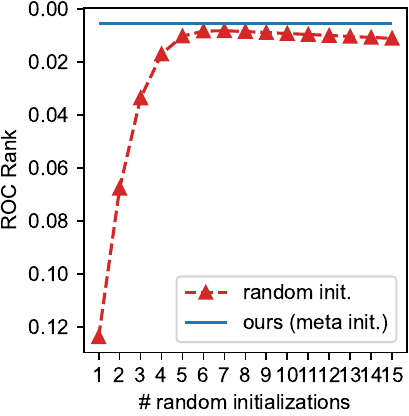}
    }
    \subfloat %
    {
    \includegraphics[clip,width=0.45\columnwidth]{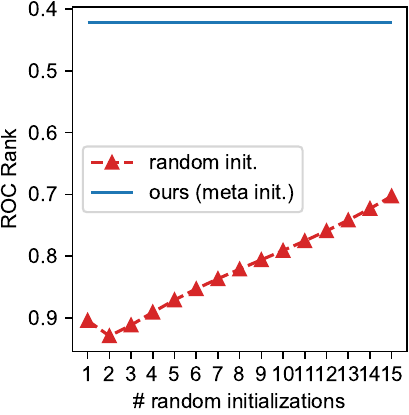}
    }
     \vspace{-0.05in}
    \caption{Comparison of ROC Rank (lower is better) of \method with meta-initialization (in blue) with increasing numbers of randomly initialized HNs, on \texttt{ODDS\_wine} (upper left), \texttt{WDBC} (upper right), \texttt{HeartDisease} (lower left) and \texttt{Ionosphere} (lower right). It needs 9 randomly initialized HNs to achieve the same performance as \method on \texttt{ODDS\_wine} . In general,\method finds a good model with much less running time.
 }
    \label{fig:init_ablations}
     \vspace{0.1in}
\end{figure}

\textbf{Effect of Meta-initialization.} In Fig. \ref{fig:init_ablations}, we demonstrate the effectiveness of meta-initialization by comparing it with random initialization on four datasets. In addition to utilizing meta-initialization, one could run \method multiple times with randomly initialized HPs and select the best model based on \f. To simulate this scenario, we vary the number of random initializations (x-axis) and record all the \f values along with the corresponding ROC Rank. For each dataset, we select the best model based on \f across \textit{all} trials. We increase the number of random trials from 1 to 15, where the highest \f value among the 15 random initialized trials is chosen as the best model. Meta-initialization is indeed a strong starting point for \method's HP tuning. For example, on the \texttt{ODDS\_wine} dataset, it requires 9 randomly initialized HNs to attain the same performance as our approach with meta-initialization, showing a 9-fold increase in the time required for online selection. 
In other cases, training 15 randomly initialized HNs fails to achieve the same performance as meta-initialization, further validating its advantages.

\begin{figure}[!th]
\centering
\vspace{-0.15in}
{\hspace{-0.13in}\subfloat
        {%
			\includegraphics[clip,width=0.38\columnwidth]{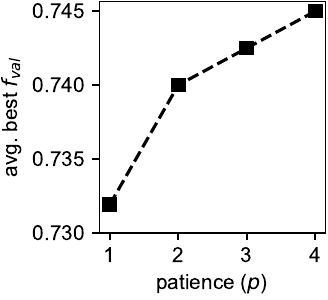}%
	}
	\hspace{0.05in}
        {%
			\includegraphics[clip,width=0.4\columnwidth]{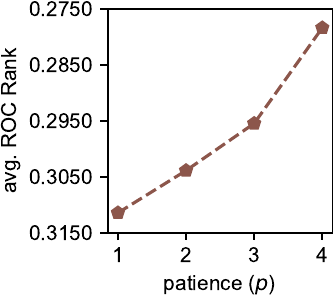}%
	}
}
\vspace{-0.1in}
\caption{Analysis of the effect of patience $p$: (left) avg. \f value change when increasing $p$ from 1 to 4; (right) avg. ROC Rank (lower is better) with increasing $p$. Larger $p$ leads to more exploration and tends to offer better performance. } \label{fig:ablation_f_roc}
\end{figure}

\textbf{Effect of Patience.} The convergence criterion for \method is based on the highest predicted performance by \f remaining unchanged for $p$ consecutive iterations (``patience''). As illustrated in Fig. \ref{fig:ablation_f_roc}, increasing the value of $p$ allows for more exploration and potentially better performance. However, this also prolongs the convergence time. In our experiments, we set $p=3$ to balance performance and runtime. The specific value of $p$ can be determined through cross-validation over the historical datasets.

\subsection{Limitation}
To analyze \method’s performance with respect to train/test datasets, we compare \method’s predictions to the test dataset’s similarity to the training datasets. We first adapt the code from MetaOD's feature extractor\footnote{\url{https://github.com/yzhao062/MetaOD/blob/master/metaod/models/gen_meta_features.py}} and extract features that represent the underlying data distribution, including mean, standard deviation, kurtosis, sparsity, skewness, and etc. Since each dataset is now represented as a vector of meta-features, we are able to measure the pairwise cosine similarity between datasets. For each dataset, we then calculate the average cosine similarity to the training datasets.

Table \ref{tab:performance_metrics} shows the Top-5 datasets where \method has the most AUROC performance differences to the Top-1 baseline (listed in Table \ref{table:full_roc}), and the dataset’s average cosine similarity to the training datasets. We observe that \method’s performance can be subpar when the test dataset has a small cosine similarity to the training datasets, since all the 5 datasets have smaller cosine similarity than the average pairwise dataset similarity. We can further conclude that one working assumption of \method is that test dataset has a similar data distribution to at least a few of the training datasets.
\vspace{-0.15in}
\begin{table}[ht]
\centering
\caption{Average Cosine Similarity to Training Dataset vs. \method's AUROC Difference. \method performs worse when test dataset has small consine similarity to training data. }
\vspace{-0.12in}
\label{tab:performance_metrics}
\begin{tabular}{@{}lcc@{}}
\toprule
\textbf{Dataset} & \textbf{Avg. Cos Sim.} & \textbf{AUROC Diff. (Rank)} \\ 
\midrule
DAMI\_Wilt & 0.4570 & 0.1371 (9) \\
ODDS\_vertebral & 0.2610 & 0.1262 (7) \\
DAMI\_Annthyroid & 0.1955 & 0.1236 (8) \\
DAMI\_Glass & 0.2451 & 0.0525 (7) \\
ODDS\_annthyroid & 0.2446 & 0.0339 (5) \\
\midrule
\textbf{Avg. Pairwise Sim.} & \textbf{0.5194} & \textbf{-} \\
\bottomrule
\end{tabular}
\end{table}
\vspace{-0.15in}

\section{Related Work}
\label{sec:related}

\hspace{0.25cm}\textbf{Supervised Model Selection.}
Supervised model selection leverages hold-out data with labels.
Randomized \cite{bergstra2012random}, bandit-based \cite{li2017hyperband}, and Bayesian optimization (BO) techniques \cite{shahriari2015taking} are various leading approaches. Self-tuning networks (STN) \cite{MacKayVLDG19} utilizes validation data to alternatively update the HPs in the HP space along with the corresponding model weights. Under the context of OD, recent work include AutoOD \cite{li2020autood} that focuses on neural architecture search, as well as PyODDS \cite{li2020pyodds} and TODS \cite{lai2020tods} for model selection, all of which rely 
on hold-out labeled data.
Clearly, these supervised approaches do not apply to UDOMS.

\textbf{Unsupervised Model Selection.~}
To choose OD models in an unsupervised fashion, one approach is to design  unsupervised internal evaluation metrics \cite{JCC8455,journals/corr/Goix16,MarquesCSZ20} that solely depend on input features, outlier scores, and/or the learned model parameters. However, a recent large-scale study showed that most internal metrics have limited performance in unsupervised OD model selection \cite{ma2021large}. More recent solutions leverage meta-learning that selects the model for a new dataset by using the information on similar historical datasets---SOTA methods include MetaOD \cite{zhao2021automatic} and ELECT \cite{zhao2022toward}. Their key bottleneck is efficiently training the candidate models with different HPs, which is addressed in our work.

\textbf{Hypernetworks.~} Hypernetworks (HN) have been primarily used for parameter-efficient training of large models with diverse architectures { \cite{ZhangRU19,MacKayVLDG19,BrockLRW18,knyazev2021parameter} as well as generating weights for diverse learning tasks 
\cite{hypermaml2022,ohs2019hypercl}}. HN generates weights (i.e. parameters) for another larger network (called the main network) \cite{HaDL17}. 
As such, one can think of the HN
as a model compression  tool for training, one that requires fewer learnable parameters. Going back in history, hypernetworks can be seen as the birth-child of the ``fast-weights'' concept by \citet{schmidhuber1992learning}, where one network produces {context-dependent weight changes} for another network.    
The {context}, in our as well as several other work \cite{BrockLRW18,MacKayVLDG19}, is the {hyperparameters} (HPs).
That is, we train a HN model that takes (encoding of) the HPs of the (main) DOD model as input, and produces {HP-dependent} weight changes for the DOD model that we aim to tune.
Training a {single} HN that can generate weights for the (main) DOD model for {varying HPs} can effectively bypass the cost of fully-training those candidate models from scratch.

\section{Conclusion}
\label{sec:conclusion}
We introduced \method, a new framework for 
unsupervised deep outlier model selection. 
\method tackles two fundamental challenges that arise in this setting: validation in the absence of supervision and  efficient search of the large model space.
To that end, it employs meta-learning to train a proxy validation function on historical datasets to effectively predict model performance on a new task without labels. 
To speed up search, it utilizes a novel hypernetwork design that generates weights for the detection model with varying HPs including model architecture,  achieving significant efficiency gains over individually training the candidate models.
Extensive experiments on a large testbed with 35 benchmark datasets showed that \method significantly outperforms 8 simple to SOTA baselines.
We expect that our work will help practitioners   use existing deep OD models more effectively as well as foster further work on unsupervised model selection in the era of deep learning.

\clearpage
\bibliographystyle{ACM-Reference-Format}
\bibliography{refs}

\appendix
\section*{Appendix}
\setcounter{table}{0}

\renewcommand{\thetable}{\Alph{section}\arabic{table}}
\renewcommand{\thefigure}{\Alph{section}\arabic{figure}}

\begin{figure*}[!ht]
\label{fig:HN-conv}
  \centering
    \includegraphics[width=0.7\textwidth]{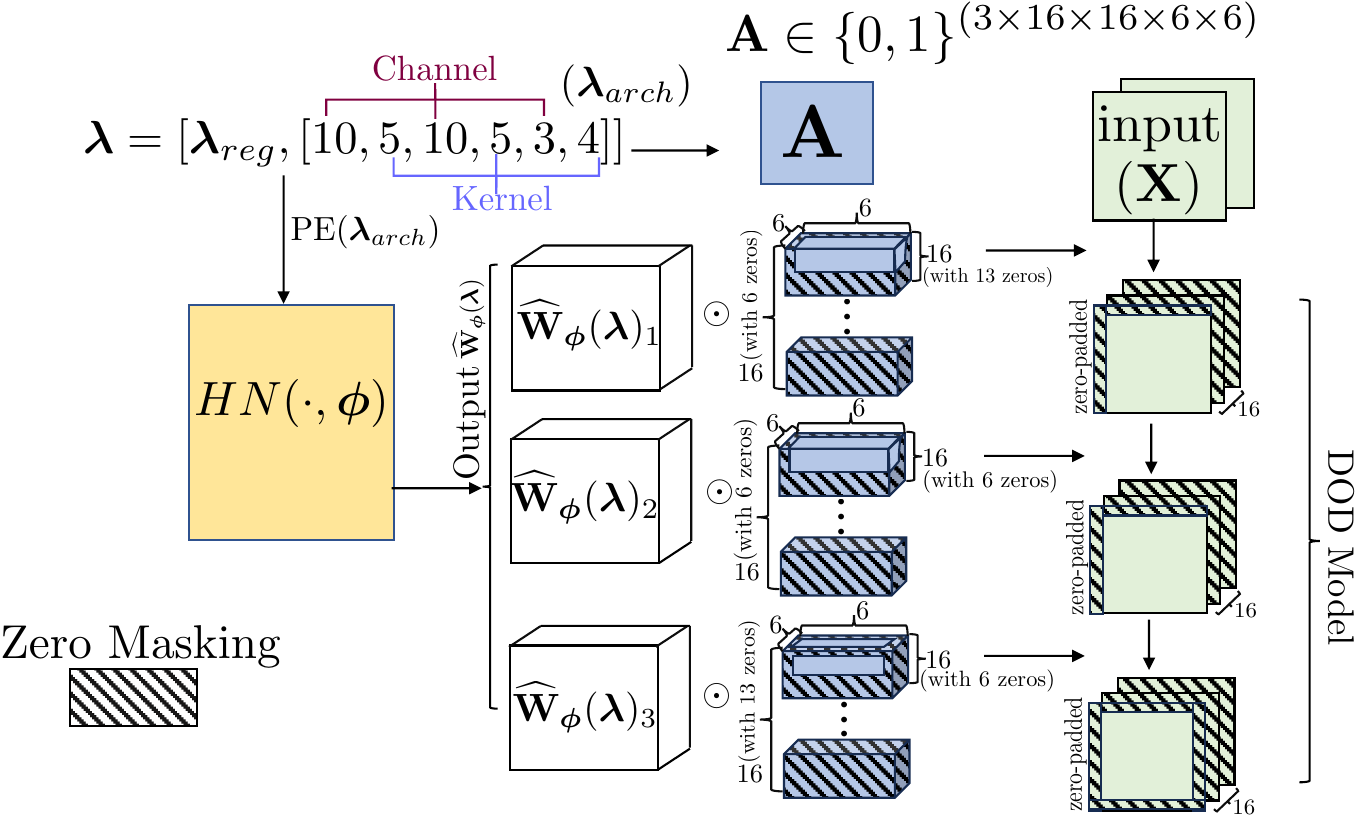}
    \vspace{-0.1in}
  \caption{Illustration of the proposed HN. HN generates weights for a 3-layer convolutional networks , with channels equal to $[10,10,3]$, and kernels equal to $[5,5,4]$.
  The HN weights $\hn$ is of size $3 \times 16 \times 16 \times 6 \times 6$, and similarily we construct the same-size architecture masking $\mathbf{A}$. At the first layer, we need to pad $\mathbf{A}$ for 1 zero, among the third and fourth dimension (we pad starting from the left and from the top). This will enable us to extend $\hn$ to a convolutional operation of kernel size $5$, from fixed kernel size $6$. To match the padding operation, we also pad the input $\mathbf{X}$ along the first and second dimension, with 1. The rest layers follow similairly. }
\end{figure*}

\section{Additional Experiment Settings and Results}

\subsection{Algorithm Settings and Baselines}
\label{appx:setting_baselines}

\textbf{Setting of HN}: The HN utilized in the experiments consists of two hidden layers, each containing 200 neurons. It is configured with a learning rate of 1e-4, a dropout rate of 0.2, and a batch size of 512. We find this setting give enough capacity to generate various weights for linearAEs. Because of the meta-learning setting, the hyperparameters of HN can be tested with validation data and test results, on historical data.

\textbf{Meta-training for \f}. Table \ref{tab:hps} includes the HP search space for training in fully-connected AE. In the table, compression rate refers to how many of the widths to shrink between two adjacent layers. For example, if the first layer has width of $6$, compression\_rate equals $2$ would gvie the next layer width equal to $3$. We also notice that some datasets may have smaller numbers of features. Thus, with the corresponding compression rate, we also have discretized the width to the nearest integer number. Thus, for some datasets, the HP search space will be smaller than $240$. In addition, for HPs in Convolutional AE, \method conducts a search among the following HPs: Number of encoders is within $[2,3,4]$, kernel size is in $[3,4,5]$, channels is $[8,16,32]$, lr is $[1e^{-4},5e^{-4}]$, weight decay is $[0,1e^{-4},1e^{-5}]$, and dropout is $[0,0.1,0.2]$. 

\textbf{HN (Re-)Training during the Online Phase}: In order to facilitate effective local re-training, we set a training epoch of $T=100$ for each iteration, indicating the sampling of 100 local HPs for HN retraining. 
In Eq. (\ref{eq:true_objective}), we designate the number of sampled HPs and the sampling factor as 500, i.e., $V_\lambda=V_\sigma=500$. 

\textbf{Specification of $\vsigma$}: It is noted that for some values of $\vsigma$, the sampled $\vlambda$ may not be a valid HP configuration. For example, it is not possible to have floating number as number of layers, or it is not practical when dropout is larger than $0.5$. We discard the impossible $\vlambda$.

\textbf{Convergence}: To achieve favorable performance within a reasonable timeframe, we set the patience value as $p=3$.

\textbf{Baselines}: We have incorporated 8 baselines, encompassing a spectrum from simple to state-of-the-art (SOTA) approaches. Table \ref{table:baseline} offers a comprehensive conceptual comparison of these baselines.

(\textit{i}) \textbf{\textit{no model selection}}:
\begin{enumerate}[label={(\arabic*)},leftmargin=0.9cm]

\setlength\itemsep{0.005in}
    \item \textbf{Default} employs the default HPs utilized in the widely-used OD library PyOD \cite{zhao2019pyod}. This serves as the default option for practitioners when no additional information is available.
    \item \textbf{Random} randomly selects an HP/model (the reported performance represents the expected value obtained by averaging across all DOD models).
\end{enumerate}

(\textit{ii}) \textbf{\textit{model selection without meta-learning}}:
\begin{enumerate}[label={(\arabic*)},leftmargin=0.9cm]
\setlength\itemsep{0.005in}
\setcounter{enumi}{2}
    \item \textbf{MC} \cite{ma2021large} utilizes the consensus among a group of DOD models to assess the performance of a model. A model is considered superior if its outputs are closer to the consensus of the group. MC necessitates the construction of a model group during the testing phase. For more details, please refer to a recent survey \cite{ma2021comprehensive}.
\end{enumerate}

(\textit{iii}) \textbf{\textit{model selection by meta-learning}} requires first 
building a corpus of historical datasets on a group of defined DOD models 
and then selecting the best from the model set at the test time. Although these baselines utilize meta-learning, none of them take advantage of the HN for acceleration.
\vspace{-0.05in}
\begin{enumerate}[label={(\arabic*)},leftmargin=0.9cm]
\setlength\itemsep{0.005in}
\setcounter{enumi}{3}
\item \textbf{Global Best (GB)} selects the best-performing model based on the average performance across historical datasets. 
\item \textbf{ISAC \cite{conf/ecai/KadiogluMST10}} groups historical datasets into clusters and predicts the cluster of the test data, subsequently outputting the best model from the corresponding cluster.
\item \textbf{ARGOSMART (AS) \cite{nikolic2013simple}} measures the similarity between the test dataset and all historical datasets, and then outputs the best model from the most similar historical dataset.
\item \textbf{MetaOD} \cite{zhao2021automatic} employs matrix factorization to capture both dataset similarity and model similarity, representing one of the state-of-the-art methods for unsupervised OD model selection.
\item \textbf{ELECT} \cite{zhao2022toward} iteratively identifies the best model for the test dataset based on performance similarity to the historical dataset. Unlike the above meta-learning approaches, ELECT requires model building during the testing phase to compute performance-based similarity.
\end{enumerate}

\textbf{Baseline Model Set}. We use the same HP search spaces for baseline models as well as the HN-trained models. Table \ref{tab:hps} provides the detailed HP search space for fully connected AE. For Conv AE, the HP search space is: Number of encoders is within $[2,3,4]$, kernel size is in $[3,4,5]$, channels is $[8,16,32]$, lr is $[1e^{-4},5e^{-4}]$, weight decay is $[0,1e^{-4},1e^{-5}]$, and dropout is $[0,0.1,0.2]$.

\subsection{Additional Results}
\label{appx:full_results}
In addition to the distribution plot in Fig. \ref{fig:perf_ae_barplot}, we provide the $p$-values of Wilcoxon signed rank test between \method and baselines in \ref{table:ae_pairs}. See \S \ref{sec:experiments} for the  experiment analysis. Full results are in Table \ref{table:full_roc}.

\begin{table}[H]
\vspace{-0.1in}
\caption{Pairwise statistical tests between \method and baselines by Wilcoxon signed rank test. \method are statistically better than baselines at different significance levels.
 }
 \vspace{-0.18in}
\centering
\scalebox{1}{
    \begin{tabular}{ll|l}
\toprule
\textbf{Ours} & \textbf{Baseline} & \textbf{p-value} \\
\midrule
\textbf{Ours}        & Default           & 0.0035           \\
\textbf{Ours}        & Random            & 0.0003         \\
\textbf{Ours}        & ISAC              & 0.0042           \\
\textbf{Ours}        & AS                & 0.0081           \\
\textbf{Ours}        & MetaOD            & 0.0051           \\
\textbf{Ours}        & Global Best       & 0.0021           \\
\textbf{Ours}        & MC                & 0.0008           \\
\textbf{Ours}        & ELECT             & 0.0803           \\
\midrule
\textbf{Ours}        & Ours (reg\&width) & 0.0001           \\
\textbf{Ours}        & Ours (reg\&depth) & 0.0001           \\
\textbf{Ours}        & Ours (reg only)   & 0.0001           \\
\bottomrule
\end{tabular}
}
	
	\label{table:ae_pairs} %
	\vspace{-0.15in}
\end{table}
\setlength\tabcolsep{6 pt}

\begin{table}[H]
	\setlength{\tabcolsep}{1pt}
\caption{
Hyperparameter search space for both free-range and HN models. We give the list of HPs as well as the range of the selected HPs.
\vspace{-0.18in}
\label{tab:hps}
}
\centering
\scalebox{0.95}{
\begin{tabular}{l|r}
\toprule
\textbf{List of Hyperparameters (HPs)}  & \textbf{\# HPs}   \\
\midrule
  n\_layers: [2,4,6,8] & 4 \\
  compression\_rate: [1.0,1.2,1.4,1.6,1.8,2.0,2.2,2.4,2.6,2.8,3.0]  & 10  \\
  dropout: [0.0,0.2,0.4] & 3 \\
  weight\_decay: [0.0,1e-6,1e-5] & 3 \\
  \midrule
  Total Number: &240 \\
 \bottomrule
\end{tabular}
}
\vspace{-0.1in}
\end{table}

\begin{table*}[htbp]
\centering
\caption{ROC and rank of the evaluated methods. The best method per dataset (row) is highlighted in bold.}
\vspace{-0.19in}
\label{table:full_roc}
\scalebox{0.815}{
\begin{tabular}{l|cccccccc|c}
\toprule
\textbf{Dataset} & \textbf{Default} & \textbf{Random} & \textbf{MC} & \textbf{GB} & \textbf{ISAC} & \textbf{AS} & \textbf{MetaOD} & \textbf{ELECT} & \textbf{Ours} \\
\midrule
\textbf{DAMI\_Annthyroid} & \textbf{0.7124 (1)} & 0.5972 (6) & 0.6123 (3) & 0.5929 (7) & 0.6018 (5) & 0.5873 (9) & 0.6050 (4) & 0.6148 (2) & 0.5888 (8) \\
\textbf{DAMI\_Cardiotocography} & 0.7159 (6) & 0.7202 (5) & 0.7024 (7) & 0.7740 (3) & 0.7571 (4) & 0.6940 (8) & 0.6458 (9) & 0.7818 (2) & \textbf{0.7866 (1)} \\
\textbf{DAMI\_Glass} & \textbf{0.7442 (1)} & 0.7055 (6) & 0.6304 (9) & 0.7244 (3) & 0.6699 (8) & 0.7230 (4) & 0.7225 (5) & 0.7431 (2) & 0.6917 (7) \\
\textbf{DAMI\_HeartDisease} & 0.3045 (9) & 0.4276 (6) & 0.4214 (7) & 0.5250 (5) & 0.5382 (2) & 0.4214 (7) & 0.5312 (4) & 0.5348 (3) & \textbf{0.5926 (1)} \\
\textbf{DAMI\_PageBlocks} & 0.8722 (8) & 0.9107 (5) & 0.9219 (2) & 0.9162 (4) & \textbf{0.9255 (1)} & 0.9002 (6) & 0.6247 (9) & 0.8791 (7) & 0.9215 (3) \\
\textbf{DAMI\_PenDigits} & 0.3837 (9) & 0.5248 (6) & 0.5422 (5) & 0.5491 (4) & 0.5069 (8) & \textbf{0.6953 (1)} & 0.6278 (3) & 0.5084 (7) & 0.6792 (2) \\
\textbf{DAMI\_Shuttle} & 0.6453 (8) & 0.9462 (2) & 0.9400 (5) & 0.9342 (7) & 0.9436 (3) & \textbf{0.9530 (1)} & 0.5525 (9) & 0.9405 (4) & 0.9391 (6) \\
\textbf{DAMI\_SpamBase} & 0.5208 (7) & 0.5232 (5) & 0.4907 (9) & 0.5210 (6) & 0.5263 (4) & \textbf{0.5552 (1)} & 0.5307 (3) & 0.5135 (8) & 0.5525 (2) \\
\textbf{DAMI\_Stamps} & 0.8687 (6) & 0.8687 (6) & 0.8926 (4)& 0.8981 (3)& \textbf{0.9079 (1)} & 0.8618 (8) & 0.7112 (9) & 0.8897 (5) & 0.9003 (2)\\
\textbf{DAMI\_Waveform} & 0.6810 (7) & 0.6772 (8) & 0.6560 (9) & 0.6941 (2) & 0.6924 (4) & 0.6890 (6) & 0.6900 (5) & \textbf{0.7019 (1) } & 0.6929 (3) \\
\textbf{DAMI\_WBC} & 0.7493 (9) & 0.9769 (6) & 0.9770 (5) & 0.9682 (8) & 0.9742 (7) & 0.9779 (3) & 0.9809 (2) & 0.9779 (3) & \textbf{0.9826 (1)} \\
\textbf{DAMI\_WDBC} & 0.8092 (7)& 0.8366 (4) & 0.8146 (9) & 0.8597 (3) & 0.8683 (2) & 0.8092 (7) & 0.8361 (5) & 0.8213 (6) & \textbf{0.9039 (1)} \\
\textbf{DAMI\_Wilt} & \textbf{0.5080 (1)} & 0.4524 (8) & 0.4832 (2) & 0.4653 (7) & 0.4700 (4) & 0.4700 (4) & 0.4714 (3) & 0.4700 (4) & 0.3709 (9)\\
\textbf{DAMI\_WPBC} & 0.4090 (8) & 0.4464 (5) & 0.3972 (9) & 0.4679 (3) & 0.4548 (4) & 0.4285 (7) & 0.4456 (6) & 0.4726 (2) & \textbf{0.4824 (1) } \\
\textbf{ODDS\_annthyroid} & \textbf{0.7353 (1)} & 0.6981 (7) & 0.6963 (8) & 0.6982 (6) & 0.7067 (2) & 0.7067 (2) & 0.6903 (9) & 0.7058 (4) & 0.7014 (5) \\
\textbf{ODDS\_arrhythmia} & 0.7769 (9) & 0.7786 (7) & 0.7810 (4) & 0.7767 (9)& \textbf{0.7831 (1)} & 0.7798 (6) & 0.7824 (3) & 0.7807 (5) & 0.7827 (2) \\
\textbf{ODDS\_breastw} & 0.5437 (9) & 0.6187 (7) & 0.8939 (3) & \textbf{0.9071 (1)} & 0.8032 (5) & 0.7986 (6) & 0.5913 (8) & 0.8649 (4) & 0.9045 (2) \\
\textbf{ODDS\_glass} & \textbf{0.6195 (1)} & 0.5849 (3) & 0.5453 (8) & 0.5897 (2) & 0.5962 (4) & 0.5962 (4) & 0.5654 (7) & 0.5957 (5) & 0.5993 (6)\\
\textbf{ODDS\_ionosphere} & 0.8708 (4) & 0.8497 (8) & \textbf{0.8711 (3)} & 0.8252 (9) & 0.8422 (7) & 0.8350 (8) & 0.8727 (2) & 0.8686 (5) & 0.8509 (6) \\
\textbf{ODDS\_letter} & 0.5555 (9) & 0.5758 (8) & 0.5918 (7) & 0.6068 (6) & 0.6244 (5) & 0.6155 (6) & \textbf{0.6446 (1)} & 0.6211 (4) & 0.6102 (8) \\
\textbf{ODDS\_lympho} & 0.9096 (9) & 0.9959 (3) & 0.9988 (2) & 0.9842 (7) & 0.9929 (5) & 0.9953 (4) & 0.9971 (4) & \textbf{1.0000 (1)} & 0.9925 (6) \\
\textbf{ODDS\_mammography} & 0.5287 (9) & 0.7612 (3) & 0.7233 (6) & 0.8362 (2) & 0.7189 (7) & 0.7116 (8) & \textbf{0.8640 (1)} & 0.7673 (4) & 0.8542 (5) \\
\textbf{ODDS\_mnist} & 0.8518 (7) & 0.8915 (4) & 0.8662 (6) & 0.8959 (3) & 0.9011 (2) & 0.8580 (5) & \textbf{0.9070 (1)} & 0.9032 (2) & 0.8994 (4)\\
\textbf{ODDS\_musk} & 0.9940 (9) & \textbf{1.0000 (1)} & \textbf{1.0000 (1)} & \textbf{1.0000 (1)} & \textbf{1.0000 (1)} & \textbf{1.0000 (1)} & \textbf{1.0000 (1)} & \textbf{1.0000 (1)} & \textbf{1.0000 (1)} \\
\textbf{ODDS\_optdigits} & 0.5104 (5) & 0.4950 (9) & 0.5092 (6) & 0.4806 (9) & 0.5115 (4) & 0.5171 (3) & 0.4973 (8) & 0.5338 (2) & \textbf{0.5584 (1)} \\
\textbf{ODDS\_pendigits} & 0.9263 (8) & 0.9295 (6) & 0.9265 (7) & 0.9305 (5) & 0.9208 (9) & 0.9386 (2) & 0.9360 (3) & 0.9346 (4) & \textbf{0.9435 (1)} \\
\textbf{ODDS\_satellite} & \textbf{0.7681 (1)} & 0.7284 (9) & 0.7445 (4) & 0.7352 (8) & 0.7433 (5) & 0.7324 (7) & 0.7486 (3) & 0.7571 (2) & 0.7432 (6) \\
\textbf{ODDS\_satimage-2} & 0.9707 (7) & 0.9826 (3) & \textbf{0.9865 (1)} & 0.9744 (6) & 0.9838 (2) & 0.9798 (5) & 0.9871 (1) & 0.9786 (8) & 0.9853 (4) \\
\textbf{ODDS\_speech} & 0.4761 (4) & 0.4756 (5) & 0.4692 (7) & 0.4726 (6) & \textbf{0.4832 (1)} & 0.4692 (7) & 0.4706 (8) & 0.4774 (3) & 0.4707 (2) \\
\textbf{ODDS\_thyroid} & \textbf{0.9835 (1)} & 0.9661 (2) & 0.9652 (3) & 0.9535 (5) & 0.9635 (4) & 0.9652 (3) & 0.9740 (2) & 0.9689 (6) & 0.9667 (7) \\
\textbf{ODDS\_vertebral} & \textbf{0.6019 (1)} & 0.5378 (2) & 0.5629 (3) & 0.5253 (4) & 0.4602 (6) & 0.5629 (3) & 0.4657 (5) & 0.5629 (3) & 0.4757 (7) \\
\textbf{ODDS\_vowels} & 0.4897 (8) & 0.5903 (7) & 0.5965 (6) & 0.6309 (5) & 0.6414 (4) & \textbf{0.6686 (1)} & 0.6216 (3) & 0.5247 (9) & \textbf{0.6686 (1)} \\
\textbf{ODDS\_wbc} & 0.4146 (9) & 0.8401 (5) & 0.7640 (7) & 0.8808 (4) & 0.8745 (6) & 0.8469 (8) & 0.8770 (3) & 0.8469 (8) & \textbf{0.9289 (1)} \\
\textbf{ODDS\_wine} & 0.7864 (2) & 0.5430 (7) & 0.4084 (8) & 0.7539 (3) & 0.5387 (6) & 0.4084 (8) & 0.6296 (4) & 0.6218 (5) & \textbf{0.8287 (1)} \\
\bottomrule
\end{tabular}
}
\end{table*}

\end{document}